\PassOptionsToPackage{table,xcdraw}{xcolor}
\documentclass[runningheads]{llncs}

\usepackage{eccv}

\usepackage[dvipsnames,table,xcdraw]{xcolor}
\usepackage{booktabs}
\usepackage{multirow}
\usepackage{pifont}
\usepackage{makecell}
\usepackage{wrapfig}
\usepackage{ulem}
\usepackage{marvosym}

\newcommand{\name} {MM-Interleaved}

\newcommand{\prompt}[1]{\textcolor{black}{\footnotesize{\texttt{#1}}}}

\definecolor{mygray}{gray}{.98}

\definecolor{mybrown}{RGB}{244,177,131}
\definecolor{mygreen}{RGB}{169,209,142}
\definecolor{myred}{RGB}{255,105,105}
\definecolor{myblue}{RGB}{157,195,230}

\usepackage{eccvabbrv}

\usepackage{graphicx}
\usepackage{booktabs}

\usepackage[accsupp]{axessibility}  % Improves PDF readability for those with disabilities.

\usepackage[pagebackref,breaklinks,colorlinks,citecolor=eccvblue]{hyperref}
\usepackage{orcidlink}

\usepackage{enumitem}

\newcommand\blfootnote[1]{%
\begingroup
\renewcommand\thefootnote{}\footnote{#1}%
\addtocounter{footnote}{-1}%
\endgroup
}

\begin{document}

% ---------------------------------------------------------------
% TODO REVIEW: Replace with your title
\title{MM-Interleaved: Interleaved Image-Text Generation via Multi-modal Feature Synchronizer} 

% TODO REVIEW: If the paper title is too long for the running head, you can set
% an abbreviated paper title here. If not, comment out.
\titlerunning{MM-Interleaved}

% TODO FINAL: Replace with your author list. 
% Include the authors' OCRID for the camera-ready version, if at all possible.
\author{
Changyao Tian$^{2,1*\dag}$,
    Xizhou Zhu$^{3,4,1*}$,
    Yuwen Xiong$^{5,1*}$,
    Weiyun Wang$^{6,1\dag}$, \\
    Zhe Chen$^{7,1\dag}$, 
    Wenhai Wang$^{2,1}$, 
    Yuntao Chen$^{8}$,
    Lewei Lu$^{4}$,
    Tong Lu$^{7}$, \\
    Jie Zhou$^{3}$,
    Hongsheng Li$^{2}$,
    Yu Qiao$^{1}$,
    Jifeng Dai$^{3,1}$\textsuperscript{\Letter}
}

% TODO FINAL: Replace with an abbreviated list of authors.
\authorrunning{C.~Tian et al.}
% First names are abbreviated in the running head.
% If there are more than two authors, 'et al.' is used.

% TODO FINAL: Replace with your institution list.
\institute{
$^1$OpenGVLab, Shanghai AI Laboratory~~~
    $^2$MMLab, CUHK~~~ \\
    $^3$Tsinghua University~~~
    $^4$SenseTime Research~~~
    $^5$University of Toronto\\
    $^6$Fudan University~~~
    $^7$Nanjing University~~~
    $^8$CAIR, HKISI, CAS~~~\\
	{\small \url{https://github.com/OpenGVLab/MM-Interleaved}} \\
}

\maketitle

\vspace{-1em}

\begin{abstract}

\blfootnote{* Equal contribution; \Letter~Corresponding author (daijifeng@tsinghua.edu.cn).}
\blfootnote{\dag~This work is done when Changyao Tian, Weiyun Wang, and Zhe Chen are interns at
Shanghai AI Laboratory.}

Developing generative models for interleaved image-text data holds both research and practical value. It necessitates models to comprehend interleaved sequences and subsequently generate images and text. However, existing attempts are limited by the issue that the fixed number of visual tokens cannot efficiently capture image details, which is particularly problematic in the multi-image scenarios. To address this, this paper presents \name, an end-to-end generative model for interleaved image-text data. It introduces a multi-scale and multi-image feature synchronizer module (MMFS), enabling intermediately direct access to fine-grained image features from the previous context during the generation process. \name~is end-to-end pre-trained on both paired and interleaved image-text corpora. It is further enhanced through a supervised fine-tuning phase, wherein the model improves its ability to follow complex multi-modal instructions. Experiments demonstrate the versatility of \name~in recognizing visual details following multi-modal instructions and generating consistent images following both textual and visual conditions. Code and models are available at \url{https://github.com/OpenGVLab/MM-Interleaved}.

\end{abstract}    
\section{Introduction}
\label{sec:intro}

Interleaved image-text data like news and blogs is a sequence of multiple images interspersed with text, which is ubiquitous on the internet.
As an extension of image-text pairs widely used in previous multi-modal models~\cite{openclip, fang2022eva, sun2023evaclip, fang2023eva02, radford2021clip}, the interleaved format not only covers a broader range of data but also presents longer and more complex article structures. 
Developing multi-modal models capable of simultaneously comprehending and generating such interleaved image-text data has significant research value and practical application potential.
It expands the scope of previous multi-modal models~\cite{sun2023emu} and enables more unified processing of multi-modal data, bridging previously disparate research fields such as text-to-image generation~\cite{rombach2022stablediffusion,ramesh2021dalle} and visual question-answering~\cite{li2023blip2, 
liu2023llava}.

\begin{figure}
    \centering
    \includegraphics[width=0.77\linewidth]{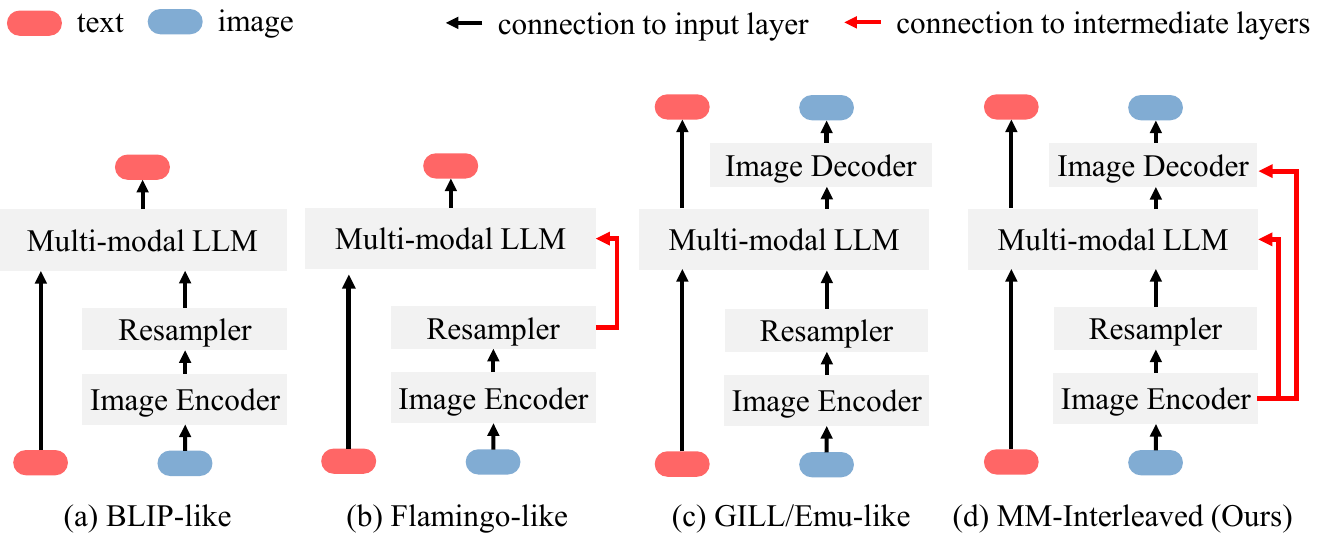}
    \vspace{-0.8em}
    \caption{\textbf{Different types of image-text generative modeling}. (a) and (b) can only generate text. (c) and (d) can generate both images and text. All types except (d) are limited by the fixed number of visual tokens extracted by context-insensitive Resampler, which will lose image details and is problematic in multi-image scenarios.}
    \vspace{-1.7em}
    \label{fig:comparison}
\end{figure}

With recent progress in multi-modal modeling with Large Language Models (LLMs)~\cite{brown2020gpt3,openai2023gpt4}, exploring how LLMs can be leveraged for interleaved image-text modeling has become increasingly attractive. A core challenge of interleaved multi-modal LLMs is how to effectively handle multiple images within the limited context length of LLMs. In most multi-modal LLMs, images are encoded as visual tokens from an image tokenizer (also known as image encoder) and fed into LLMs together with text tokens~\cite{koh2023gill, dong2023dreamllm, sun2023emu, jin2023unified}.
However, due to computational and memory constraints, multi-modal LLMs have limited context window sizes (\eg, 2048 or 4096 tokens).
To reduce computational demands and the required context length of LLMs, Perceiver Resamplers~\cite{alayrac2022flamingo} are commonly used to map each image from up to 1024 tokens to a small fixed number of visual tokens (\eg, 32 or 64 tokens) as in Fig.~\ref{fig:comparison}.

Due to the relatively small visual token number, critical image details may be overlooked, especially in tasks requiring fine-grained observation. 
Although increasing the number of visual tokens~\cite{lv2023kosmos2_5, yu2023cm3leon} may address this issue to some extent, the token number per image must be restricted to ensure complete input into the LLMs. Allocating a large number (\eg, 2890 in~\cite{tian2022sphinx}) of visual tokens per image would pose a significant limitation especially in multi-image scenarios.

\begin{figure}[t]
    \centering
    \includegraphics[width=0.7\linewidth]
    {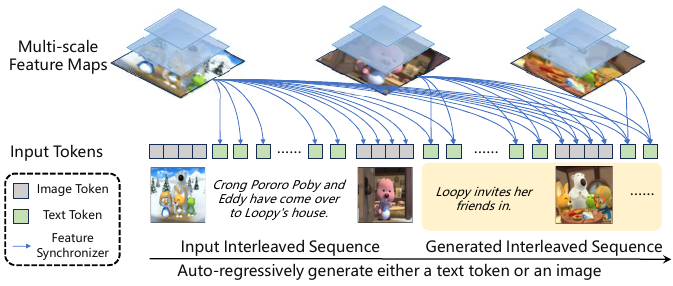}
    \vspace{-0.7em}
    \caption{\textbf{Illustration of Multi-Modal Feature Synchronizer (MMFS).}
    In the auto-regressive generation of interleaved image-text sequences, besides interacting with low-resolution image tokens via self-attention, tokens in \name{} could also use MMFS to cross-attend multi-scale high-resolution image features. 
    MMFS ensures the attention will have exactly the same causal relation between images and text tokens.
    }
    \vspace{-2.0em}
    \label{fig:sync}
\end{figure}

We note that the main problem here is the context insensitivity of Perceiver Resamplers, as it only takes image features as input. Each input image is compressed into a fixed number of tokens, making it challenging or even infeasible to preserve all requisite information and accommodate subsequent generation requirements.
However, such information loss can actually be mitigated through further feature extraction in the intermediate layers of the LLMs. Based on the intermediate features of the LLMs, relevant information can be dynamically extracted from the original images to fulfill the perceptual requirements.
Such a dynamic extraction mechanism can also help to effectively handle scenarios involving multiple images and high-resolution image feature maps.

To facilitate dynamic image feature extraction in the intermediate layers of the LLMs, we design a fine-grained Multi-Modal Feature Synchronizer (MMFS) based on the Deformable Sparse Attention~\cite{zhu2020deformable}, which can efficiently extract relevant information from multi-scale image feature maps and multiple images as shown in Fig.~\ref{fig:sync}.
Based on MMFS, we further propose \name{}, a novel end-to-end generation model for processing interleaved image-text data, as shown in Fig.~\ref{fig:framework}, which can generate both textual and visual response given interleaved image-text inputs.

Specifically, input images and text are first mapped into tokens through their respective tokenizers (\ie, image encoder and word embedding) and then fed to the LLM, arranged in their original order. 
A special token  \verb|<BoI>| is introduced to represent ``Begin of Image''. When the LLM processes the input sequences, each token in the intermediate layers directly observes multi-image and multi-scale feature maps from the previous context through MMFS. After being processed by the LLM, the features of the text tokens are then used to predict the next text word. When the \verb|<BoI>| token is predicted, an image decoder based on diffusion models is used to predict the next image. All previous LLM output features are passed into the image decoder as the generation conditions. With MMFS, the image decoder can also extract details of previous images intermediately.

\begin{figure}[t]
    \centering
    \includegraphics[width=0.93\linewidth]{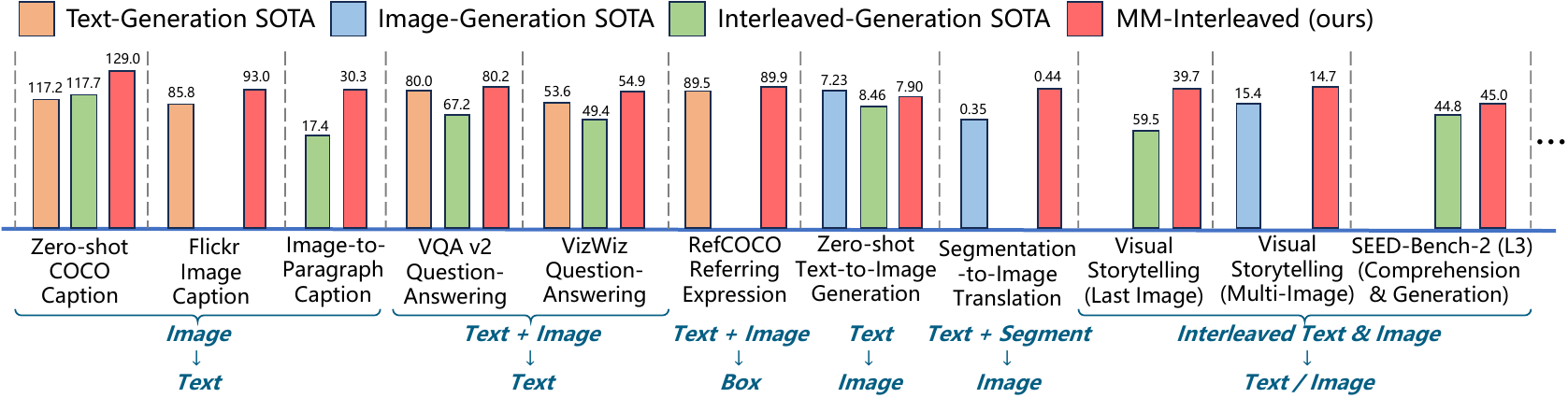}
    \vspace{-0.8em}
    \caption{\textbf{Comparison on multi-modal benchmarks.} 11 representative tasks are listed. See \cref{sec:exp} for other benchmarks. \name{} (\textcolor{myred}{red}) consistently outperforms previous methods supporting interleaved generation (\textcolor{mygreen}{green}). It also surpasses or is on par with specialists in either text generation (\textcolor{mybrown}{brown}) or image generation (\textcolor{myblue}{blue}).}
    \vspace{-1.3em}
    \label{fig:performance}
\end{figure}

\name{} is pre-trained on a mixture of image-text pairs and interleaved image-text sequences without using any in-house data. 
Similar to previous multi-modal LLMs, supervised fine-tuning can further enhance the model capabilities. Due to the end-to-end generative modeling, fine-tuning can be applied to both text and image generation. Our model is fine-tuned on several tasks, including visual question-answering, image captioning, referring expression grounding, text-to-image generation, segmentation-to-image translation, and visual storytelling. As is illustrated in \cref{fig:performance}, our model achieves competitive results and shows higher token efficiency compared with previous image-to-text and text-to-image methods. Compared with joint image and text generation models, we set the new SOTA results for a wide range of tasks.
The key contributions are:
\vspace{-0.5em}
\begin{itemize}[leftmargin=*]
    \item We propose Multi-Modal Feature Synchronizer (MMFS) to reduce the number of visual tokens required by multi-modal LLMs, which can efficiently extract fine-grained visual details from multi-scale feature maps of multiple images according to the intermediate context features of the multi-modal LLMs;
    \item We propose \name{} based on MMFS for generative modeling of interleaved image-text data, which can preserve fine-grained image information with only a small number of visual tokens per image, and can be optimized end-to-end for both text and image generation;
    \item Our method can generate both accurate text descriptions and visually consistent images given interleaved image-text inputs, achieving SOTA results on various multi-modal comprehension and generation benchmarks without using any in-house data.
\end{itemize}

\section{Related Work}
\label{sec:related}

\noindent\textbf{Modeling of Paired Image-Text Data} plays an important role in the progress of multi-modal research in recent years.
 A series of public large-scale image-text pairs datasets have been released \cite{openclip,gadre2023datacomp,schuhmann2022laion5b}. Based on these data, models trained by image-text contrastive learning~\cite{radford2021clip,jia2021scaling,sun2023evaclip,li2023flip} are able to recognize and understand open-world semantics. Subsequent works~\cite{wang2021simvlm,zhu2021uni,li2021albef,yu2022coca} further incorporated text generation tasks such as image captioning, while other works~\cite{rombach2022stablediffusion,ramesh2021dalle} were proposed to generate images based on text prompts.
The latest progress of LLMs~\cite{openai2023gpt4,openai2022chatgpt} has 
launched a new era, leading to the emergence of many LLM-based 
multi-modal models trained using image-text pairs~\cite{li2023blip2, 
liu2023llava, wang2023visionllm, wang2023allseeing}. For example, LLaVA~\cite{liu2023llava} connects LLMs with pretrained visual encoders 
using linear projections to build powerful multi-modal foundation models.

\vspace{0.3em}\noindent\textbf{Modeling of Interleaved Image-Text Data} has received increasing attention recently. Early works such as Kosmos~\cite{huang2023kosmos1} and Flamingo~\cite{alayrac2022flamingo} focused on understanding such data with non-public datasets. To promote the development of this field, public and large-scale interleaved image-text datasets were later released in~\cite{zhu2023mmc4,laurenccon2023obelics}.
More recent works~\cite{zhao2023mmicl,zhang2023internlmxcomposer} have concentrated on understanding the interleaved data. Nevertheless, their generative capabilities remain confined to text. Initial attempts at generating images and text given interleaved contexts were first undertaken in~\cite{koh2023gill, yu2023cm3leon, sun2023emu, dong2023dreamllm}. A two-stage generation process was introduced by \cite{koh2023gill, sun2023emu}, wherein text and images are generated in the first and second stages, respectively. CM3Leon~\cite{yu2023cm3leon} employs VQ-VAE~\cite{van2017vqvae} to convert images into discrete tokens, facilitating token-level auto-regressive modeling as language modeling. However, it primarily focuses on image generation capabilities and exhibits notable weaknesses in image understanding. DreamLLM~\cite{dong2023dreamllm} focuses on single-stage end-to-end modeling using raw image pixels as inputs. 
Emu2~\cite{sun2023emu2}, SEED-LLaMA~\cite{ge2023making} and VL-GPT~\cite{zhu2023vlgpt} adopt additional training stage for the image tokenizer-detokenizer. 
Despite these efforts, they only feed image information at the LLM inputs, which are limited by the problem that fixed number of visual tokens cannot efficiently preserve image details.

\vspace{0.3em}\noindent\textbf{Integrating Image Details into LLMs} is important for multi-modal LLMs. Most works use Perceiver Resamplers~\cite{alayrac2022flamingo,li2023blip2,zhu2023minigpt4} to extract image information via cross-attention, mapping each image into a fixed number of visual tokens. For example, Flamingo~\cite{alayrac2022flamingo} adopts Resamplers in the intermediate layers of LLMs, injecting extracted image features into LLMs through gated residual connections. BLIP-2\cite{li2023blip2} and Mini-GPT4~\cite{zhu2023minigpt4} introduce Resamplers at the bottom of LLM to insert the extracted visual tokens into the input text sequences.
While these methods achieve good performance, image details remain to be overlooked due to the small number (\eg, 32 or 64) of visual tokens.
To preserve more details, recent works~\cite{bai2023qwenvl, lv2023kosmos2_5, liu2023llava, chen2023shikra} increase the number of input visual tokens per image to hundreds. In SPHINX~\cite{tian2022sphinx}, the token number is further increased to 2,890. Although it helps mitigate information loss, the computational and memory demands of LLMs are also significantly increased. 
Increasing the number of visual tokens per image is particularly problematic in multi-image scenarios, where multiple images naturally require more visual tokens, making it hard to use for interleaved image-text data.

\vspace{-0.5em}
\section{Method}
\label{sec:method}

\subsection{Task Formulation}
\label{subsec:formu}

To build an end-to-end generative model for interleaved image-text data, 
we first consider an interleaved image-text sequence $X=\{x_1, x_2, x_3, \dots\}$, where each element $x_n$ is either a text token (denoted as $x_n^L$) or a whole image (denoted as $x_n^V$). Text and images are arranged in the order in which they appear in the original content. 
In multi-modal LLMs, a common practice is to first extract embeddings for each text token and each image and then feed them into LLMs, \ie, $e_n^L = \mathcal{E}_L(x_n^L)$ and  $e_n^V = \mathcal{E}_V(x_n^V)$, where $\mathcal{E}_L$ denotes word embedding following standard practices in NLP. $\mathcal{E}_V$ is typically an image encoder (\eg, ViTs~\cite{dosovitskiy2020image}) followed by a Perceiver Resampler~\cite{alayrac2022flamingo} to map each image to a fixed number of tokens.
Then, the generative modeling is trained to maximize the log-likelihood:
\begin{equation}
\footnotesize
    \begin{aligned}
        \log p(X) = \sum_n \log p(x_n | e_{< n})
        = \sum_{n \in \mathcal{I}_L} \log p(x_n^L | e_{< n}) + \sum_{n \in \mathcal{I}_V} \log p(x_n^V | e_{< n}),
    \end{aligned}
    \label{equ:formu}
\end{equation}
where $\mathcal{I}_L$ and $\mathcal{I}_V$ represent the index sets for text tokens and images, respectively. $<n$ in the subscript represents the abbreviation of $\{1, 2, \dots, n-1\}$. The following paragraphs provide explanations of Eq.~\eqref{equ:formu}.

\begin{figure*}[t]
    \centering
    \includegraphics[width=0.9\linewidth]{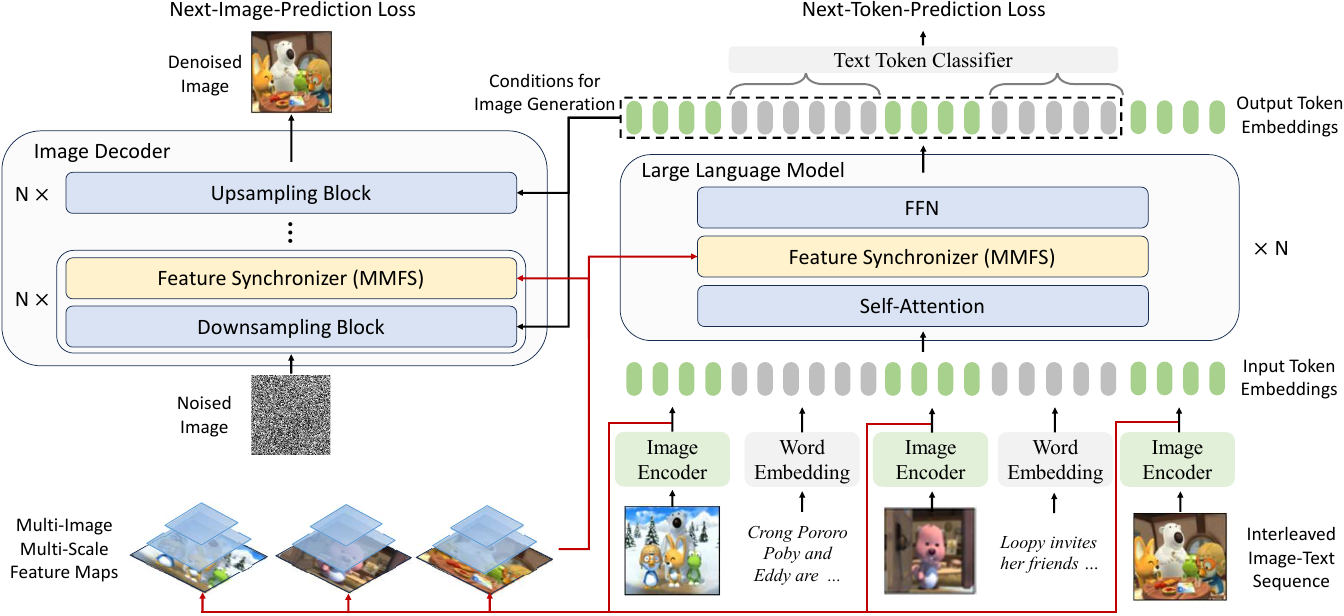}
    \vspace{-0.7em}
    \caption{
        \textbf{Architecture of {\name}}. The red lines represent how the multi-scale image features are generated and utilized. {\name} incorporates image encoder to both extract high-resolution multi-scale image features (red lines) and map each image into a fixed number of low resolution visual tokens. These visual tokens are fed into the multi-modal LLM along with text tokens. LLM then uses our proposed feature synchronizer (MMFS) to extract additional high-resolution image details, and auto-regressively generate text tokens. After that, a diffusion-based image decoder generates next image conditioned on the previous context features from LLM, where MMFS is also utilized to capture accurate visual conditions.
    }
    \vspace{-2.0em}
    \label{fig:framework}
\end{figure*}

\vspace{0.3em}\noindent\textbf{Text Generation with Multi-modal Condition.}
$\log p(x_n^L | e_{< n})$ is similar to traditional causal language modeling, except that the condition also includes previous images. 
Recent works~\cite{li2023blip2, liu2023llava, wang2023visionllm, wang2023allseeing} have demonstrated the effectiveness of using LLMs for processing additional visual inputs. They apply a linear classifier on top of the LLMs. 
The loss function for text generation is
\begin{equation}
	\small
    \begin{aligned}
        L_\text{NTP}(x_n^L | e_{< n}) = - \bar{x}_n^L \cdot \log \mathrm{softmax} \big(W \cdot \mathcal{D}_\text{LLM}(e_{< n}) \big),
    \end{aligned}
    \label{equ:loss_text}
\end{equation}
where $W$ is the linear classification weight, $\mathcal{D}_\text{LLM}$ denotes the LLM network (\eg, LLaMA~\cite{touvron2023llama}), $\bar{x}_n^L$ is the one-hot vector indicating the ground-truth text.

\vspace{0.3em}\noindent\textbf{Image Generation with Multi-modal Condition.}
Maximizing $\log p(x_n^V | e_{< n})$ aligns with the denoising-diffusion process~\cite{luo2022understanding}, which recently achieves widespread success in image generation. Maximizing the log-likelihood is derived as minimizing the diffusion modeling loss as
\begin{equation}
	\small
    \begin{aligned}
        L_\text{NIP}(x_n^V | e_{< n}) = \mathbb{E}_{\epsilon,t}~||\epsilon - \mathcal{D}_\text{DM}\big(x_{n,t}^V, t, \mathcal{D}_\text{LLM}(e_{< n})\big)||^2,
    \end{aligned}
    \label{equ:loss_img}
\end{equation}
where $\mathcal{D}_{DM}$ is the diffusion model for image denoisiong. $x_{n,t}^V$ is the noisy version of the original image at the denoising step $t$, and the denoising network $\mathcal{D}_{DM}$ is trained to predict the noise $\epsilon$.
We found that such modeling is also applicable when conditional input is interleaved image-text data.

\vspace{0.3em}\noindent
Such modeling allows for the flexible combination of different language models, image encoders, and image decoders to fully leverage a variety of pre-trained models. 
The entire framework can be optimized end-to-end.

\subsection{Architecture}
\label{subsec:architecture}

Building upon the task formulation in Sec.~\ref{subsec:formu}, we propose a novel multi-modal architecture for processing interleaved image-text data. 
It integrates a Visual Foundation Model (VFM), a Large Language Model (LLM), and a Diffusion Model (DM).
Such integration aims to excel in both comprehension and generation tasks of text and images by leveraging the strengths of each model type.
As illustrated in Fig.~\ref{fig:framework}, our architecture comprises three key components:

(1) \textbf{VFM-based Image Tokenizer $\mathcal{E}_V$} that maps each image $x^V \in \mathbb{R}^{H \times W \times 3}$ (\eg, $H=W=224$) into a fixed number of visual tokens $e^V \in \mathbb{R}^{N \times C}$ ($N = 32$ by default). $C$ is the channel dimension. It consists of a pre-trained vision encoder (\eg, CLIP-ViT~\cite{radford2021clip}) for feature extraction and a Perceiver Resampler~\cite{alayrac2022flamingo} to reduce the number of visual tokens. 
A ViT-Adapter~\cite{chen2022vitadapter} is also used to extract multi-scale image features $F^V \in \mathbb{R}^{(\sum_{i=1}^{L} H_i \times W_i) \times C}$ for fine-grained feature fusion in subsequent networks, where $L = 3$ and $H_i = \frac{H}{2^{i+2}}, W_i = \frac{W}{2^{i+2}}$ by default. 

(2) \textbf{LLM-based Multi-modal Model $\mathcal{D}_\text{LLM}$} that extracts context features from the interleaved image-text sequences. A pre-trained LLM (\eg, Vicuna~\cite{zheng2023vicuna}) is utilized. Its input sequence $E\in \mathbb{R}^{K \times C}$ is a concatenation of embeddings $(e_1, e_2, \dots)$, where $e_n$ is either a word embedding $e_n^L \in \mathbb{R}^{1 \times C}$ or an image embedding $e_n^V \in \mathbb{R}^{N \times C}$. $K$ is the total number of input tokens. 
We also introduce a feature synchronizer (\ie, MMFS) for letting the intermediate layers in $\mathcal{D}_\text{LLM}$ directly access and extract multi-scale image features on demand.

(3) \textbf{DM-based Image Decoder $\mathcal{D}_\text{DM}$} that generates the image based on previous image-text sequences. A pre-trained diffusion model (\eg, Stable Diffusion~\cite{rombach2022stablediffusion}) is utilized. To provide the conditional inputs for $\mathcal{D}_\text{DM}$, Resampler~\cite{alayrac2022flamingo} is employed to map the output features from LLM to a fixed number of conditional tokens. The fine-grained feature synchronizer module is also used here for providing detailed visual conditions, which is very useful for tasks requiring visual alignment (\eg, image translation). 

Then we introduce the details of the proposed architecture.

\vspace{0.3em}\noindent\textbf{Multi-Modal Feature Synchronizer (MMFS).} MMFS aims to enable dynamic and efficient image detail extraction in decoder intermediate layers, compensating for information loss due to limited input visual tokens. It leverages the Deformable Attention~\cite{zhu2020deformable} to achieve efficient and sparse image attention.
MMFS can be applied to both image and text decoding, avoiding information bottlenecks caused by the Resamplers in multi-modal LLMs. It is especially efficient for processing multiple high-resolution images in context.

\begin{figure}[t]
    \centering
    \includegraphics[width=0.9\linewidth]
    {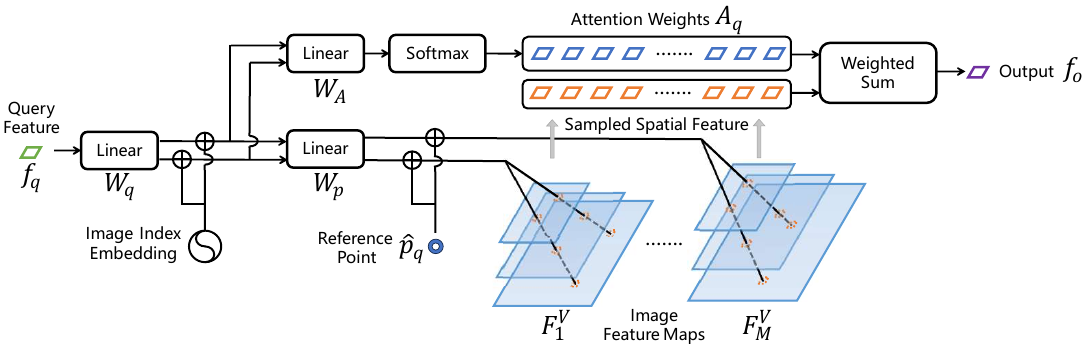}
    \vspace{-0.7em}
    \caption{
        \textbf{Architecture of MMFS module}. The query feature is passed through a linear projection and added with the image index embedding. Two linear projections are used to predict the sampling offsets and unnormalized attention weights of each image, respectively. The sampling offsets are added with the query's reference point to form the corresponding sampling locations, which are shared across all feature maps of the same image. The output is the weighted sum of the sampled spatial features. 
    }
    \vspace{-2.0em}
    \label{fig:MMFS}
\end{figure}

As is shown in Fig.~\ref{fig:MMFS}, given a query token that requires detailed image features, the MMFS module only attends to a small set of sampling points around a reference point on the reference images. Let $f_q \in \mathbb{R}^{C}$ represents the features of the query token. $\hat{p}_q \in [0,1]^2$ denotes the relative image coordinate of its reference point, which is designed for compressing the search space for sampling points from global positions to local relative offsets. By default, without a spatial prior, $\hat{p}_q = (0.5, 0.5)$ is set as the center of the images. $\{F^V_m\}_{m=1}^M$ are the multi-image multi-scale feature maps extracted by the image tokenizer, $M$ is the number of reference images. The output feature $f_o \in \mathbb{R}^C$ is given by
\begin{equation}
\footnotesize
    \begin{aligned}
        q^{(m)} &= W_{q} \cdot f_q + \text{PosEmbed}(m), \\
        p_q^{(m)} &= W_{p} \cdot q^{(m)} + \hat{p}_q, \ \ A_q^{(m)} = W_{A} \cdot q^{(m)}, \\
        p_q &= \mathrm{Concat} (p_q^{(1)}, \cdots, p_q^{(M)}), \\
        A_q &= \mathrm{softmax} \left(\mathrm{Concat}(A_q^{(1)}, \cdots, A_q^{(M)})\right), \\
        f_o &= \mathrm{DeformAttn}(\{F^V\}_{m=1}^M, A_q, p_q),
    \end{aligned}
\end{equation}
where $\text{PosEmbed} \in \mathbb{R}^{\bar{M} \times C} $ is a learnable positional embedding, $m$ indexes from the maximum of $\bar{M}$ reference images. $W_q, W_p, W_A$ are learnable linear projection weights. The coordinate of sampling points $p_q^{(m)}$ and the corresponding attention weights $A_q^{(m)}$ are first calculated for each image separately. Then, the attention weights are normalized among different images via the softmax function. The $\mathrm{DeformAttn}$ operator extracts feature at coordinates $p_q \in \mathbb{R}^{M \times L \times K \times 2 }$ from the corresponding feature map, and performs a weighted summation according to $A_q \in \mathbb{R}^{M \times L \times K}$. Here, $L$ and $K$ denote the number of multi-scale feature levels and sampling points per feature map, respectively.

\vspace{0.3em}\noindent\textbf{Multi-modal LLM with MMFS}.
The input of multi-modal LLMs is an interleaved sequence of image and text token embeddings, which always starts with a special token \verb|<s>| and ends with another special token \verb|</s>|. Image token embeddings are inserted at the corresponding positions in the original sequence. Special token \verb|<BoI>| is added in front of each image to represent ``Begin of Image''.

MMFS modules are inserted between the self-attention layer and the feed-forward layer of the LLM every fixed number of blocks. Query token $f_q$ iterates over each token in the LLM, which can only access  the previous images. $\hat{p}_q = (0.5, 0.5)$. The output of MMFS is multiplied by tanh($\alpha$) before added back to $f_q$, where $\alpha$ is a zero-initialized learnable scalar.

\vspace{0.3em}\noindent\textbf{Image Decoder with MMFS}.
The output features from the multi-modal LLM are further processed by another Resampler before fed into the image decoder. For each image to be generated, the Resampler maps previous context features to a fixed number of tokens (\eg, 77 tokens) to match the condition token length of the pre-trained diffusion model.

MMFS modules are inserted after each downsampling block in the U-Net of the diffusion model. Query token $f_q$ iterates over each pixel in the feature map. $\hat{p}_q$ is set as the spatial coordinate of the query pixel. The output of MMFS is further processed by a zero-initialized convolution before added back to $f_q$.

\vspace{0.3em}\noindent\textbf{Training Target and Inference Pipeline}. The training objective is defined as the sum of Next-Text-Token Prediction loss in Eq.~\eqref{equ:loss_text} and Next-Image Prediction loss in Eq.~\eqref{equ:loss_img} as $\mathcal{L} =  \mathcal{L}_{NTP} + \lambda~\mathcal{L}_{NIP}$,
where $\lambda$ is a hyperparameter used to determine the relative loss weight between the image and text decoding branches. The whole framework can be optimized end-to-end.

During inference, the images and texts are generated in an auto-regressive manner. Text tokens are sampled from the distribution predicted by the multi-modal LLM. When the generated token is \verb|<BoI>|, the diffusion model is called for generating the next image.

\vspace{-0.5em}
\section{Experiment}

\label{sec:exp}

\subsection{Implementation Details}

\noindent\textbf{Network.}
We adopt CLIP-ViT-L/14~\cite{radford2021clip}, Vicuna-13B~\cite{zheng2023vicuna} and Stable Diffusion v2.1~\cite{rombach2022stablediffusion} as the image encoder, large language model, and image decoder, respectively. For the multi-modal LLM, a Perceiver Resampler with 12 blocks is used to reduce the number of visual tokens per image to 64. MMFS is inserted every 4 blocks in the LLM. For the image decoder, a Perceiver Resampler with only 1 block is used to reduce the number of previous conditional tokens to 77. MMFS is inserted after each downsampling block in the image decoder. 

\vspace{0.3em}
\noindent\textbf{Pre-training.}
Our model is pre-trained on a mixture of image-text pairs and interleaved image-text sequences, including MMC4~\cite{zhu2023mmc4}, LAION-2B~\cite{schuhmann2022laion5b}, LAION-COCO~\cite{schuhmann2022laioncoco}, CC-12M~\cite{changpinyo2021cc12m} and Objects365~\cite{shao2019objects365}. 
For CC-12M~\cite{changpinyo2021cc12m} and Objects365~\cite{shao2019objects365}, instead of utilizing the original annotations, we use the pre-trained BLIP-2 model~\cite{li2023blip2} to caption the images.
The sampling probability of MMC4 is twice that of other image-text pair datasets. No in-house data is used. 

The images are inserted before or after the corresponding text sentence with equal probability. To optimize training efficiency and data utility, multiple image-text pairs or interleaved image-text sequences are concatenated into extended sequences with the maximum context length (\ie, 2048 tokens).

The model is pre-trained for 15000 steps with 11 billion tokens visited. The image encoder and LLM are frozen. 
The learning rate is set to be $10^{-5}$ for the image decoder and $10^{-4}$ for the rest trainable parameters. 
The input and output image resolutions are $224 \times 224$ and $512 \times 512$, respectively. 

\begin{table*}[t]
\centering
\footnotesize
\setlength\tabcolsep{2pt}
\resizebox{0.95\linewidth}{!}{
\begin{tabular}{@{}llccc|cccc|cccccc@{}}
\toprule

Model & LLM &H&I&A & COCO & Flickr & NoCaps & I2Para. & VQA$^{\rm v2}$ & OKVQA & GQA & VizWiz & TextVQA & VisDial  \\ \midrule
\multicolumn{15}{l}{\textit{Models for Text-Generation Only}} \\ 

MetaLM~\cite{hao2022metavlm} & MetaLM & - & - & - &
82.2 & 43.3 & 58.7 & -- &
41.1 & 11.4 & -- & \uline{41.1} & 11.4  \\

OF-9B~\cite{awadalla2023openflamingo} & MPT-7B &-&-&- &
79.5 & 59.5 & -- & -- & 
52.7 & 37.8 & -- & 27.5 & 24.2 & -- \\

IDEFICS-80B~\cite{idefics2023} & LLaMA-65B &-&-&- &
91.8 & 53.7 & 65.0 & & \uline{60.0} & -- & 45.2 & 36.0 & \uline{30.9} & --  \\

KOSMOS-1~\cite{huang2023kosmos1} & MetaLM &H&-&- & 
-- & 65.2 & -- & -- & 
46.7 & -- & -- & -- & -- & --  \\

KOSMOS-2~\cite{peng2023kosmos2} & KOSMOS-1 &H&-&- & 
-- & 66.7 & -- & -- &
45.6 & -- & -- & -- & -- & -- \\

Flamingo-9B~\cite{alayrac2022flamingo} & Chinchilla-7B &H&-&- & 
79.4 & 61.5 & -- & -- &
51.8 & 44.7 & -- & 28.8 & 31.8 & 48.0  \\

Flamingo-80B~\cite{alayrac2022flamingo} & Chinchilla-70B &H&-&- & 
84.3 & 67.2 & -- & -- &
56.3 & 50.6 & -- & 31.6 & 35.0 & 52.0  \\

IDEFICS-80B-I~\cite{idefics2023} & LLaMA-65B &-&I&- &
117.2 & 65.3 & 104.5 & & 37.4 & -- & -- & 26.0 & --&  \\

mPLUG-DocOwl~\cite{ye2023mplugdocowl} & LLaMA-7B &-&I&A & 
52.6& 62.2 & 57.4 & -- & -- & -- & -- & -- & -- & -- \\

BLIP-2~\cite{li2023blip2} & Vicuna-7B &-&I&A &
-- & 74.9 & 107.5 & -- & -- & -- & 38.6 & 25.3 & 40.1 & --  \\

BLIP-2~\cite{li2023blip2} & Vicuna-13B &-&I&A &
-- & 71.6 & 103.9 & -- & 41.0 & -- & 41.0 & 19.6 & 42.5 & --  \\

InstructBLIP~\cite{instructblip} & Vicuna-7B &-&I&A &
-- & 82.4 & \textbf{123.1} & -- & --& -- & 49.2 & 34.5 & 50.1 & -- \\

InstructBLIP~\cite{instructblip} & Vicuna-13B &-&I&A &
--& 82.8& 121.9 & -- & --& -- & 49.5& 33.4& 50.7 & -- \\

Shikra~\cite{chen2023shikra} & Vicuna-13B &-&I&A &
117.5 & 73.9 & -- & -- &77.4 & -- & -- & -- & -- & --  \\

LLaVA-1.5~\cite{liu2023improved} & Vicuna-7B &-&I&A 
 & -- & -- & -- & -- & 78.5 & -- & 62.0 & 50.0 & 58.2 & --  \\

LLaVA-1.5~\cite{liu2023improved} & Vicuna-13B &-&I&A & -- & -- & -- & -- & \textbf{80.0} & -- & \textbf{63.3} & 53.6 & \textbf{61.3} & --  \\

Qwen-VL~\cite{bai2023qwenvl} & Qwen-7B &H&I&A & --& 85.8 & 121.4 & -- & 78.8 & -- & 59.3 & 35.2 & 63.8 & -- \\

Qwen-VL-Chat~\cite{bai2023qwenvl} & Qwen-7B &H&I&A & --& 81.0 & 120.2 & -- & 78.2 & -- & 57.5& 38.9 & 61.5 & -- \\

\midrule
\multicolumn{15}{l}{\textit{Models for both Image and Text Generation}}\\ 
CM3Leon~\cite{yu2023cm3leon} & -- &H&-&- &
61.6 & -- & -- & 10.5 &
47.6 & 23.8 & -- & 37.6 & -- & 22.6 \\

Emu~\cite{sun2023emu} & Vicuna-13B &H&-&- &
112.4 & -- & -- & -- & 
52.0 & 38.2 & -- & 34.2 & -- & 47.4  \\

Emu-I~\cite{sun2023emu} & Vicuna-13B &H&-&- &
117.7 & -- & -- & -- & 
40.0 & 34.7 & -- & 35.4 & -- & 48.0  \\

Emu2~\cite{sun2023emu2} & LLaMA-33B &H&-&- &
-- & -- & -- & -- & 
33.3 & 26.7 & -- & 40.4 & 26.2 & --  \\

DreamLLM~\cite{dong2023dreamllm} & 
Vicuna-7B &-&I&- & 115.4& -- & --& 17.4& 56.6& 44.3
& -- & 38.1 & 34.9& --  \\

VL-GPT~\cite{zhu2023vlgpt} & LLaMA-7B &-&-&- &
116.4 & -- & -- & -- & 
51.7 & 35.8 & 34.6 & 34.7 & -- & 49.9  \\

VL-GPT-I~\cite{zhu2023vlgpt} & LLaMA-7B &-&I&A &
133.7 & -- & -- & -- & 
67.2 & 50.3 & 51.5 & 38.9 & -- & 51.8  \\

SEED-LLaMA~\cite{ge2023making} & LLaMA2-Chat-13B &-&I&A & 125.0 & -- & -- & -- & 48.1 & 27.1 & -- & 23.3 & -- & -- \\

SEED-LLaMA-I~\cite{ge2023making} & LLaMA2-Chat-13B &-&I&A & 126.9 & -- & -- & -- & 63.4 & 43.2 & -- & 49.4 & -- & -- \\

\rowcolor{mygray}
\name & Vicuna-13B &-&-&- & 
\uline{129.0} & \uline{85.8} & \uline{106.4} & \uline{23.5} & 57.0 & \uline{40.0} & \uline{46.7} & \uline{40.8} & \uline{37.2} & \uline{48.7} \\
\rowcolor{mygray}
\name-SFT & Vicuna-13B &-&I&A & 
\textbf{140.5} & \textbf{93.0} & \textbf{123.2} & \textbf{30.3}  & \textbf{80.2} & \textbf{51.7} & 60.5 & \textbf{54.9} & \textbf{61.0} & \textbf{53.7} \\ 
\bottomrule
    
\end{tabular}}
\vspace{0.3em}
\caption{\textbf{Multi-modal comprehension evaluation}. ``H'' denotes using in-house data, ``I'' means the training images of some benchmarks are included in the training, ``A'' means the training annotations of some benchmarks are visible in training. Benchmarks include COCO~\cite{chen2015cococaption}; Flickr: Flickr30k~\cite{plummer2015flickr30k}; NoCaps~\cite{agrawal2019nocaps}; I2Para.: Image2Paragraph~\cite{krause2017hierarchical}; VQA$^{\rm v2}$: VQAv2~\cite{goyal2017making}; OKVQA~\cite{marino2019ok}; GQA~\cite{hudson2019gqa}; VizWiz~\cite{gurari2018vizwiz}; TextVQA~\cite{singh2019towards}; VisDial~\cite{das2017visual}. The results in \uline{underline} and \textbf{bold} are the best performance without ``HIA'' data and without ``H'' data, respectively. The evaluation metrics for each benchmark are listed in Appendix. }
\vspace{-3.1em}
\label{table:pretrain}
\end{table*}

\begin{table*}[ht]
\centering
\footnotesize
\resizebox{0.65\linewidth}{!}{
\begin{tabular}{@{}llccccc@{}}
\toprule

Model & LLM & $L_1$ (Part-1) & Part-2 & $L_2$ & Part-3 & $L_3$ \\

\midrule
Emu~\cite{sun2023emu} & LLaMA-13B        & 42.5 & 41.1 & 42.4 & 41.4 & 42.3 \\
Next-GPT~\cite{wu2023next} & Vicuna-7B   & 30.7 & 35.6 & 31.1 & 33.9 & 31.4 \\
SEED-LLaMA~\cite{ge2023making}~ & LLaMA2-Chat-13B & \textbf{43.9} & 43.4 & \textbf{43.8} & \textbf{52.3} & \textbf{44.8} \\

\rowcolor{mygray}
\name{}    & Vicuna-13B   & \textbf{43.9}   & \textbf{46.1}   & \textbf{44.1}   & \textbf{52.1}    & \textbf{45.0}         \\ 

\bottomrule
\end{tabular}}
\vspace{0.3em}
\caption{\textbf{Zero-shot results for interleaved image-text comprehension and generation on SEED-Bench-2~\cite{li2023seedv2}}. The average task accuracy across corresponding evaluation dimensions is reported. 
$L_1$(part-1) evaluates the image and text comprehension, $L_2$(part-1\&2) evaluates interleaved image-text comprehension, and $L_3$(part-1\&2\&3) evaluates image and text generation.
}
\vspace{-2.0em}
\label{table:seedv2}
\end{table*}

 \begin{table}[t]
\centering
\footnotesize
\resizebox{0.45\linewidth}{!}{
\begin{tabular}{@{}lcc@{}}
\toprule
Model                      & MS-COCO    & LN-COCO   \\ \midrule
\multicolumn{3}{l}{\textit{Text-to-Image Specialists}}          \\
Retrieval Result           & 17.97      & 33.59     \\
DALL-E~\cite{ramesh2021dalle}                     & $\sim$28   & -         \\
CogView2~\cite{ding2022cogview2}                   & 24.00      & -         \\
Stable Diffusion~\cite{rombach2022stablediffusion}                     & 12.43      & 34.26     \\
GLIDE~\cite{nichol2021glide}                      & 12.24      & -         \\
Make-A-Scene~\cite{gafni2022make_a_scene}               & 11.84      & -         \\
DALL-E 2~\cite{ramesh2022dalle2}                   & 10.39      & -         \\
Muse-3B~\cite{yang2020muse}                    & 7.88       & -         \\
Imagen-3.4B~\cite{saharia2022imagen}                & 7.27       & -         \\
Parti-20B~\cite{yu2022parti}                  & 7.23       & 15.97     \\ \midrule
\multicolumn{3}{l}{\textit{Models for both Image and Text Generation}} \\
CM3-13B~\cite{aghajanyan2022cm3}                    & 29.56      & -         \\
VL-GPT~\cite{zhu2023vlgpt}              & 12.25      & -         \\
GILL~\cite{koh2023gill}                       & 12.20      & -         \\
Emu-13B~\cite{sun2023emu}                    & 11.66      & -         \\
Next-GPT~\cite{wu2023next}                   & 11.28       & -          \\
CM3Leon-7B~\cite{yu2023cm3leon}                 & 10.82      & -         \\
DreamLLM-7B-Stage1~\cite{dong2023dreamllm}         & 8.76       & 22.42     \\
DreamLLM-7B~\cite{dong2023dreamllm}                & 8.46       & 20.53     \\
\rowcolor{mygray}
\name         & 7.90       & 23.88     \\ \bottomrule
\end{tabular}
}
\vspace{0.3em}
\caption{\textbf{Zero-shot text-to-image generation results.} FID~\cite{heusel2017gans} is reported.}
\vspace{-3.5em}
\label{table:image-generation}
 \end{table}

\vspace{0.3em}\noindent\textbf{Supervised Fine-tuning.}
After pre-training, like other multi-modal LLMs, our model capabilities can be further enhanced using supervised fine-tuning. The model is end-to-end fine-tuned on four types of downstream tasks: 1) visual question-answering and image caption, 2) visual storytelling, 3) segmentation-to-image translation, and 4) referring expression comprehension.

\vspace{0.2em}\noindent More implementation details could be found in Appendix.

\vspace{-0.5em}
\subsection{Evaluation}
We evaluate the zero-shot capability of the pre-trained \name{} as well as its performance on various downstream tasks after supervised fine-tuning.

\vspace{0.3em}\noindent\textbf{Zero-shot Results after Pre-training.}
\cref{table:pretrain} demonstrates our strong performance on multi-modal zero-shot comprehension across various benchmarks, including image captioning (COCO~\cite{chen2015cococaption}, Flicker30K~\cite{plummer2015flickr30k}, NoCaps~\cite{agrawal2019nocaps}, I2Para~\cite{krause2017hierarchical}), visual question answering (VQAv2~\cite{goyal2017making}, OKVQA~\cite{marino2019ok}, VizWiz~\cite{gurari2018vizwiz}, TextVQA~\cite{singh2019towards}) and visual dialogue (VisDial~\cite{das2017visual}). 
In the fully decontaminated setting where the image and text in downstream tasks are unseen during pre-training, our model significantly outperforms other methods on all tasks. This demonstrates the effectiveness of the proposed \name{} approach. Furthermore, our model even exceeds most trained on vast in-house data like Flamingo-9B~\cite{alayrac2022flamingo} and Emu~\cite{sun2023emu}, showing the importance of proper architecture for image-text interaction.

\cref{table:image-generation} shows the results on text-to-image generation for MS-COCO~\cite{lin2014microsoft} and LN-COCO~\cite{pont2020connecting}. On MS-COCO, we sample 8 images per text condition and use CLIP~\cite{radford2021clip} to rerank based on text-image similarity. CLIP reranking is not used for LN-COCO. Our model achieves competitive performance compared to existing image and text generation models. Note that some other works (\eg, Emu, Muse, Imagen, and Parti) use in-house data, while ours did not.

\cref{table:seedv2} shows the zero-shot performance for interleaved image-text comprehension and generation on SEED-Bench-2~\cite{li2023seedv2}.
Our pre-trained model achieves SOTA results on both comprehension and generation tasks.
Visualization results for interleaved image-text generation are showed in Appendix.

\begin{table*}[t]
\centering
\footnotesize
\resizebox{0.62\linewidth}{!}{
\begin{tabular}{@{}lcccccccc@{}}
\toprule
\multicolumn{1}{c}{\multirow{2}{*}{Model}} & \multicolumn{3}{c}{RefCOCO~\cite{kazemzadeh2014refcoco}} & \multicolumn{3}{c}{RefCOCO+~\cite{mao2016refcoco_plus_g}} & \multicolumn{2}{c}{RefCOCOg~\cite{mao2016refcoco_plus_g}} \\ \cmidrule(l){2-9} 
\multicolumn{1}{c}{}                       & Val     & Test-A  & Test-B  & Val     & Test-A   & Test-B  & Val           & Test         \\ \midrule
OFA-L~\cite{wang2022ofa}                   & 79.96   & 83.67   & 76.39   & 68.29   & 76.00    & 61.75   & 67.57         & 67.50        \\
VisionLLM-H~\cite{wang2023visionllm}       & -       & 86.70   & -       & -       & -        & -       & -             & -            \\
Shikra~\cite{chen2023shikra}            & 87.01   & 90.61   & 80.24   & 81.60   & 87.36    & 72.12   & 82.27         & 82.19        \\
MiniGPT-V2~\cite{chen2023minigptv2}     & 88.69   & 91.65   & 85.33   & 79.97   & 85.12    & 74.45   & 84.44         & 84.66        \\
Ferret~\cite{you2023ferret}            & \textbf{89.48}   & \textbf{92.41}   & 84.36   & \textbf{82.81}   & \textbf{88.14}    & 75.17   & \textbf{85.83}         & \textbf{86.34}        \\
*~Qwen-VL~\cite{bai2023qwenvl}            & \textbf{89.36}   & \textbf{92.26}   & 85.34   & \textbf{83.12}   & \textbf{88.25}    & \textbf{77.21}   & \textbf{85.58}         & 85.48        \\
\rowcolor{mygray}
\name{}                   & \textbf{89.92}   & \textbf{92.59}   & \textbf{86.54}   & \textbf{82.99}   & \textbf{88.57}    & \textbf{77.07}   & \textbf{85.21}         & 84.92        \\ \bottomrule
\end{tabular}
}
\vspace{0.5em}
\caption{\textbf{Supervised fine-tuning results on referring expression comprehension task}. * denotes using an additional self-constructed grounding dataset and trained with an image resolution larger than 224.}
\vspace{-2.1em}
\label{table:grounding}
\end{table*}

\vspace{0.3em}\noindent\textbf{Fine-tuning Results.}
\cref{table:pretrain} also shows our fine-tuned results (MM-Interleaved-SFT) on multi-modal comprehension. Our fine-tuned model achieves SOTA performance without using any in-house data.
On visual question answering tasks, it matches the previous best LLaVA-1.5 model.
Compared to LLaVA-1.5, our model has two key advantages: 1) our method is capable of generating both images and text, while LLaVA-1.5 can only generate text; 2) LLaVA-1.5 uses 576 visual tokens as LLM inputs, whereas we only require 64 tokens. Our model achieves competitive image understanding with far fewer visual tokens, making it better suited for multi-image scenarios.

\cref{table:grounding} shows the results on referring expression comprehension (REC) benchmarks. Our method outperforms other methods. 
Even though we only use public REC data~\cite{kazemzadeh2014refcoco,mao2016refcoco_plus_g} for fine-tuning, \name{} matches Qwen-VL~\cite{bai2023qwenvl} which utilizes an extra 22M in-house grounding dataset and trains at higher resolution (448 vs. 224 pixels used by ours).
This shows the effectiveness of synchronized fine-grained image features for enhancing the REC capability of our method. 

\cref{table:sft-ade20k} shows the performance on segmentation-to-image translation, which is a capability not shown in other multi-modal LLMs.
We follow the protocol proposed in ControlNet~\cite{zhang2023adding} by generating images from ADE20K ground truth semantic masks and use OneFormer~\cite{jain2023oneformer} for segmenting the generated images. Standard mIoU metric is used for measuring the semantic and structural fidelity of the generated images.
\name{} clearly outperforms other baselines including ControlNet by a large margin. 
Compared to ControlNet, \name{} could leverage the better representations learned by our generative modeling from large-scale pre-training data, which benefits the image translation tasks.
Moreover, thanks to MMFS, \name{} is capable of generating realistic images with pixel-level precise alignment from a semantic label map, where simply relying on coarse Perceiver Resampler features fails on this task (see \cref{tab:ade_ablate}).

\begin{table*}[t]
\footnotesize
\label{table:sft}
\begin{subtable}[b]{0.34\linewidth}
\centering
\setlength{\tabcolsep}{2pt}
\resizebox{0.95\linewidth}{!}{
\begin{tabular}{ccc}
\toprule
\textit{Groundtruth} & VQGAN~\cite{esser2021taming} &  LDM~\cite{rombach2022high} \\ 
\textit{0.58}  & 0.21 &  0.31  \\
\midrule
PIPT~\cite{wang2022pretraining} & ControlNet~\cite{zhang2023adding} & Ours\\
 0.26  & 0.35 & {\bf 0.44}\\
\bottomrule
\end{tabular}}
\vspace{0.5em}
\caption{
Segmentation-to-image generation on ADE20K~\cite{zhou2017ade20k}. mIoU is reported.
}
\label{table:sft-ade20k}
\end{subtable}
\hfill
\begin{subtable}[b]{0.31\linewidth}
\centering
\setlength{\tabcolsep}{2pt}
\resizebox{0.9\linewidth}{!}{
\begin{tabular}{@{}lcc@{}}
\toprule
Model & CLIP Sim.$\uparrow$ & FID$\downarrow$ \\
\midrule
GILL~\cite{koh2023gill} & 0.64 & - \\
MiniGPT-5~\cite{zheng2023minigpt5} & \textbf{0.70}&59.5\\
\rowcolor{mygray}
\name & \textbf{0.70} & \textbf{39.7} \\ \bottomrule
\end{tabular}}
\vspace{0.5em}
\caption{Last image generation with interleaved context for visual storytelling on VIST~\cite{huang2016vist}.}
\label{table:sft_vist}
\end{subtable}
\hfill
\begin{subtable}[b]{0.32\linewidth}
\centering
\setlength{\tabcolsep}{2pt}
\resizebox{0.9\linewidth}{!}{
\begin{tabular}{@{}lcc@{}}
\toprule
Model  &  Pororo      & Flintstones \\ \midrule
StoryDALL-E~\cite{maharana2022storydall_e}  &      25.9    &  26.5 \\
AR-LDM~\cite{pan2022ar_ldm}  &      17.4         &  19.3 \\
ACM-VSG~\cite{feng2023acm_vsg}  &     15.4         &  \textbf{18.4} \\
\rowcolor{mygray}
 \name  &  \textbf{14.7}          & \textbf{18.7} \\
\bottomrule
\end{tabular}}
\caption{Multi-image generation with interleaved context for visual storytelling. FID is reported.}
\label{table:sft_pororo}
\end{subtable}

\vspace{-0.8em}
\caption{\textbf{Segmentation-to-image generation and visual storytelling results.}}
\vspace{-3.1em}
\end{table*}

\cref{table:sft_vist,table:sft_pororo} show our results on visual storytelling, which is required to generate new images with interleaved image-text as context.
We evaluate the performance both for generating the last image on the VIST dataset~\cite{huang2016vist} and for generating multiple images auto-regressively on the Pororo~\cite{li2019pororo} and Flintstones~\cite{gupta2018flintstones} datasets.
Our model achieves SOTA performance even compared with previous specialist methods. Qualitative results can be found in Appendix.

\vspace{-0.5em}
\subsection{Ablation Study}

For ablation study, we use CLIP-ViT-L/14 \cite{radford2021clip} as the image encoder, OpenLLaMA-3B v2~\cite{openlm2023openllama} as the LLM and miniSD\footnote{\scriptsize https://huggingface.co/justinpinkney/miniSD} as the image decoder. 
The output image resolution is 256. 
The model is pre-trained for 10000 steps, using a mixture of LAION-COCO~\cite{schuhmann2022laioncoco}, LAION-2B~\cite{schuhmann2022laion5b}, and MMC4~\cite{zhu2023mmc4} datasets. 

For simplicity, MMFS in this subsection is by default implemented in a single-image and single-scale setting, where each token only attends to the $16\times16$ feature map of the nearest preceding image.
We evaluate the zero-shot performance on three representative tasks and four datasets, \ie, image captioning on the COCO Karpathy test set, text-to-image generation on the COCO Karpathy test set, visual question answering on OKVQA~\cite{marino2019ok}, and TextVQA~\cite{singh2019towards} validation set. 
The evaluation metrics are CIDEr, FID-5k, and top-1 accuracy, respectively. 
To ablate the cross-attention module in image generation, we also evaluate the fine-tuned results for segmentation-to-image translation task on ADE20K~\cite{zhou2017ade20k}.

\setlength{\tabcolsep}{2pt}
\begin{table*}[t]
\footnotesize

\begin{subtable}[b]{0.42\linewidth}
\centering
\resizebox{\linewidth}{!}{
\begin{tabular}{@{}cc|ccccc@{}}
\toprule
\# Token  &  w/ MMFS  & Caption$\uparrow$ & Generation$\downarrow$ & OK-VQA$\uparrow$ & TextVQA$\uparrow$\\ \midrule
 32  &                &   107.0       & 32.2      &  28.7  &   22.5  \\
\rowcolor{mygray}
 32  &   \checkmark   &   \textbf{110.6}       & \textbf{30.0}      &  \textbf{29.8}  &   \textbf{27.7} \\
256  &                &   \textbf{110.7}       & 32.7      &  29.2  &   23.6  \\
256  &   \checkmark   &   \textbf{110.9}       & \textbf{29.5}      &  \textbf{29.6}  &  \textbf{27.8}  \\
\bottomrule
\end{tabular}}
\caption{pre-training with 224 resolution}
\label{table:ablate_token}
\end{subtable}
\hfill
\begin{subtable}[b]{0.38\linewidth}
\centering
\resizebox{0.98\linewidth}{!}{
\begin{tabular}{@{}c|cccc@{}}
\toprule
w/ MMFS & Caption$\uparrow$ & Generation$\downarrow$ & OK-VQA$\uparrow$ & TextVQA$\uparrow$ \\ \midrule
                    &   110.5       & \textbf{30.3}      &  29.9  &   24.9  \\
\rowcolor{mygray} 
\checkmark          &   \textbf{115.2}       & \textbf{30.5}      &  \textbf{30.6}  &   \textbf{30.8}  \\
\bottomrule
\end{tabular}}
\vspace{0.5em}
\caption{fine-tuning with 448 resolution}
\label{table:ablate_res448}
\end{subtable}
\hfill
\begin{subtable}[b]{0.18\linewidth}
\centering
\resizebox{0.85\linewidth}{!}{
\begin{tabular}{c|c}
\toprule
 w/ MMFS  &  ADE20k$\uparrow$ \\
\midrule
           & 5.3 \\
\rowcolor{mygray} 
\checkmark & \textbf{35.9} \\
\bottomrule
\end{tabular}}
\caption{fine-tuning for image translation}
\label{tab:ade_ablate}
\end{subtable}
\vspace{-1.8em}
\caption{\textbf{Ablation on using MMFS}. ``Generation'' is text-to-image generation task. ``ADE20k'' is segmentation-to-image translation task. Others are text generation tasks. ``\# Token'' indicates the number of input visual tokens for LLMs (32 by default).}
\vspace{-2.0em}
\end{table*}
\begin{table}[t]
\footnotesize
\centering
\resizebox{0.75\linewidth}{!}{
\begin{tabular}{@{}ccc|cccc@{}}
\toprule
Cross-Attn         &  Transition & Attn Input & COCO Cap.$\uparrow$ & COCO Gen.$\downarrow$  & OK-VQA$\uparrow$  & TextVQA$\uparrow$  \\ \midrule
\rowcolor{mygray}
Deformable        &  None             & $16\times16$ &   \textbf{110.6}       & \textbf{30.0}      &  \textbf{29.8}  &  \textbf{27.7}  \\
Dense             & None        & $16\times16$        &   108.5       & 30.6      &  28.4  &   23.6  \\
Dense             & Resampler   & 32 tokens &   107.2       & 30.7      &  28.9 &   24.0  \\
\bottomrule
\end{tabular}
}
\vspace{0.5em}
\caption{\textbf{Ablation on the design choice of MMFS}. Different attention modules can be used in MMFS. We also ablate whether to add additional transition layer before feeding image features into MMFS. Dense cross-attention with Resampler transition on single scale and single image is similar to the cross attention used in Flamingo~\cite{alayrac2022flamingo}.}
\vspace{-3.3em}
\label{table:ablate_resampler}
\end{table}

\vspace{0.3em}\noindent\textbf{Token Efficiency.} 
As shown in Tab.~\ref{table:ablate_token}, when equipped with MMFS, using only 32 visual tokens per image can outperform using 256 visual tokens without MMFS. Such results demonstrate the effectiveness of our method when the context length is limited. 
As shown in Tab.~\ref{table:ablate_res448}, the performance improvement of using MMFS becomes larger when increasing the input image resolution from 224 to 448.
Such results indicate our method could better exploit the additional information gained from high resolution even with only 32 visual tokens.

\vspace{0.3em}\noindent\textbf{MMFS for Image Generation.} 
Tab.~\ref{tab:ade_ablate} demonstrates the criticality of adding MMFS for segmentation-to-image translation task.
This task is hard as it requires precise pixel-level information to align the given segmentation condition and image output properly.
Without using MMFS, this task fails, showing extremely low mIoU results. 
Visualizations results in Appendix shows that the generated image without MMFS cannot preserve all spatial information, and the spatial alignment for the generated results is poor.

\vspace{0.3em}\noindent\textbf{Comparison between Different Cross Attention Mechanisms.} 
The ablation of adopting different cross-attention mechanisms for LLM is shown in Tab.~\ref{table:ablate_resampler}. 
The overall performance drops when directly replacing MMFS with the vanilla dense cross-attention, possibly due to its slower convergence speed.
We highlight that the model with Deformable attention performs significantly better than other attention mechanisms on TextVQA, indicating that Deformable attention could effectively and efficiently capture fine-grained information like text needed for the task, such as visual question answering in this case.

\begin{figure}[t]
\centering
\includegraphics[width=0.9\linewidth]{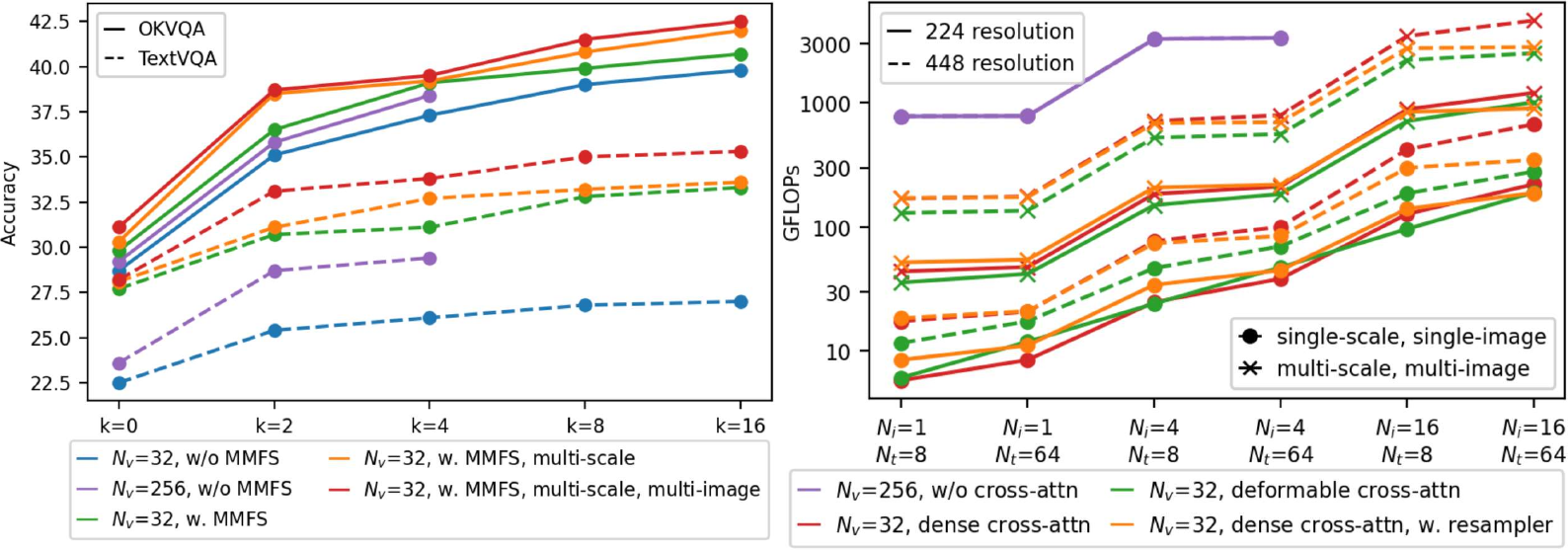}
\vspace{-0.8em}
\caption{
    \textit{Left}: \textbf{Few-shot results on {\footnotesize OKVQA} and {\footnotesize TextVQA}}. \textit{Right}: \textbf{Additional GFLOPs over using 32 visual tokens with different numbers of image and text inputs}. $N_v$ and $k$ denote the number of visual-token per image and the number of few-shot examples, respectively. 
    $N_i$ and $N_t$ denote the number of image and the number of subsequent text tokens for each image in the sequence.
     Note that $N_v=256$ does not support $N_i\geq$ 8 images due to 2048 LLM context length.
}
\label{fig:flops_fewshot}
\vspace{-2.0em}
\end{figure}

\vspace{0.3em}\noindent{\textbf{MMFS with Multi-Image and Multi-Scale}}. As shown in \cref{table:supp_msmi}, adding multi-image and multi-scale for MMFS improves the performance. To better take advantage of the multi-image mechanism, we further evaluate our model with few-shot prompts following~\cite{alayrac2022flamingo, sun2023emu}. 
As is illustrated in \cref{fig:flops_fewshot}~(left), when the number of context images increases, MMFS continuously outperforms only using 32 visual tokens, and also benefits from further attending to the multi-scale feature maps of multiple images.

\begin{table}[t]
\footnotesize
\centering
\setlength\tabcolsep{2pt}
 \resizebox{0.6\linewidth}{!}{
\begin{tabular}{@{}cc|cccc@{}}
\toprule

 multi-scale & multi-image   & Caption$\uparrow$ & Generation$\downarrow$ & OK-VQA$\uparrow$ & TextVQA$\uparrow$ \\  \midrule
             &               &   110.6       & 30.0      &  29.8  &   27.7 \\

     \ding{51}        &               &    111.2      &   29.5    &  30.3  &  28.1  \\

     \ding{51}        &      \ding{51}         &   111.2    &  29.9   &  31.1     & 28.2   \\

\midrule
\multicolumn{6}{l}{\textit{finetuning with 448 input resolution}} \\

             &               &    115.2      &  30.5     &  30.6  &   30.8 \\

     \ding{51}        &               &   115.4       &   30.1    &  31.0  &  31.3  \\

     \ding{51}        &      \ding{51}         &   115.8       &   30.0    & 31.7  & 32.0   \\

\bottomrule
\end{tabular}
 }
 \vspace{0.3em}
\caption{\textbf{Ablation on multi-scale and multi-image usage of MMFS}.}
\label{table:supp_msmi}
\vspace{-2.0em}
\end{table}

\vspace{0.3em}\noindent{\textbf{Computational Efficiency}}. 
Compared to only using 32 visual tokens, further integrating MMFS into LLM only increases by about 2\%, 2\%, 6\% and 3\% in FLOPs, \#parameter, runtime and memory, respectively. 
When compared to using more visual tokens (\ie, 256) w/o MMFS, using MMFS with 32 visual tokens achieve better performance (see \cref{table:ablate_token}) and is much more efficient, with 2.8$\times$ fewer FLOPs and 1.3$\times$ fewer runtime.
\cref{fig:flops_fewshot}~(right) shows the additional FLOPs over only using 32 visual tokens with synthesized interleaved image-text inputs. Our method consistently achieves lower or similar computation than using 256 visual tokens or using dense cross-attention.

\vspace{-0.5em}
\section{Conclusion}
\label{sec:conclusion}

This paper introduces \name{}, an end-to-end trained generative model for interleaved image-text comprehension and generation.
It is built upon the Multi-Modal Feature Synchronizer (MMFS), which is proposed to enhance multi-modal LLMs by reducing the required visual tokens, enabling efficient extraction of visual details.
Without using in-house data, our method achieves SOTA performance on various multi-modal benchmarks. 
We hope this work can contribute to the research on multi-modal LLMs for interleaved image-text scenarios.

\vspace{0.1em}\noindent{\textbf{Limitations}}. 
Both the quality and quantity of public interleaved image-text data  are relatively low, making it difficult to exploit the full potential of interleaved generative modeling.

\vspace{0.1em}\noindent{\textbf{Negative impacts}}. As other multi-modal models, it may suffer from hallucination issues and potentially generate biased contents due to the noisy training data. 
We hope to invest more efforts to improve the quality and quantity of interleaved image-text data, with the aim of further improvements while ensuring its safety and reliability.

% \clearpage  % TODO REVIEW/FINAL: This \clearpage needs to be removed from both review and camera-ready versions.

% WARNING: do not forget to delete the supplementary pages from your submission 

% ---- Bibliography ----
%
% BibTeX users should specify bibliography style 'splncs04'.
% References will then be sorted and formatted in the correct style.
%
\bibliographystyle{splncs04}
\bibliography{main}

\clearpage
\appendix

\section{Implementation Details}

\subsection{Pre-training}

\noindent\textbf{Dataset Details}. We use MMC4~\cite{zhu2023mmc4}, LAION-2B, LAION-COCO~\cite{schuhmann2022laioncoco}, CC-12M~\cite{changpinyo2021cc12m} and Objects365~\cite{shao2019objects365} as the pre-training dataset. LAION-2B is the English subset of LAION-5B~\cite{schuhmann2022laion5b}, and we further filter it based on additional metrics including aesthetics scores. LAION-COCO ~\cite{schuhmann2022laioncoco} is a 600M subset of LAION-2B, which is captioned by pre-trained BLIP~\cite{li2022blip} models. Text prompts with length shorter than 10 are also filtered out. For CC-12M~\cite{changpinyo2021cc12m} and Objects365~\cite{shao2019objects365}, instead of utilizing the original annotations, we use the pre-trained BLIP-2 model~\cite{li2023blip2} to caption the images. Following previous works~\cite{dong2023dreamllm, sun2023emu}, additional filtering rules are appled to the MMC4 dataset~\cite{zhu2023mmc4}. Specifically, images with a CLIP similarity score below 0.24 will be discarded, and only 6 images at most will be kept for each document. We also exclude 100\% of all documents that do not contain any images, and 50\% of documents that contain only 1 image.

\vspace{0.3em}
\noindent\textbf{Data Concatenation Strategy}. For image-text-pair datasets (\ie, LAION-2B, LAION-COCO~\cite{schuhmann2022laioncoco}), we randomly sample multiple image-text pairs from the same dataset and concatenate them to the maximum context length (\ie, 2048) during pre-training. For interleaved image and text datasets (\ie, MMC4~\cite{zhu2023mmc4}), we also split and concatenate the documents to form the training samples. Such concatenation strategy can utilize the full context window of Large Language Models and thus achieve high data efficiency.

\vspace{0.3em}
\noindent\textbf{More Training Details.}
The detailed hyper-parameters of pre-training are listed in \cref{tab:hp_pretrain}. Besides that, for image generation, we ignore the training loss of images that are the first element in the sequence. The text condition of the rest images are dropped with a 10\% probability to improve classifier-free guidance sampling~\cite{ho2022classifier}. 

\begin{table}[h]
    \centering
    \small
    \setlength\tabcolsep{1.0pt}
    \footnotesize
    \resizebox{0.95\linewidth}{!}{
        \begin{tabular}{lc}
        \toprule
            Hyper-parameters & Value \\
        \midrule
            Input image resolution & $224\times 224$ \\
            Output image resolution & $512\times 512$ \\
    
            VFM & CLIP-ViT-L/14 (frozen) \\
            LLM & Vicuna-13B v1.3 (frozen) \\
            DM & Stable Diffusion v2.1 \\
            
            $\lambda$ & 10 \\
            Cross-attention frequency & 4 \\
            LLM Multi-scale feature maps & $i=2,3,4$ \\
    
            Learning rate & \makecell{1e-5 (image decoder) \\ 1e-4 (others)} \\
            Weight decay & 0.05 \\
            Warmup steps & 1k \\
            Learning rate schedule & constant with warmup \\
            
            Training iterations & 15k \\
            Context length & 2048 \\
            
            Optimizer & AdamW \\
            Optimizer hyper-parameters & $\beta_1, \beta_2, \epsilon$ = 0.9, 0.995, 1e-6 \\
            
            Data & MMC4, LAION-2B, LAION-COCO, CC-12M, Objects365  \\
            Augmentation & CenterCrop \\
            Batch size (per GPU) & 2 for MMC4, and 4 for the rest \\

        \bottomrule
        \end{tabular}
    }
    \vspace{0.3em}
    \caption{Hyper-parameters for pre-training.}
    \vspace{-1em}
    \label{tab:hp_pretrain}
\end{table}

\subsection{Supervised Fine-tuning}

\noindent\textbf{VQA and Image Captioning.}
For this task, we train our model in the form of question answering, \ie, using \prompt{Based on the image, please answer the question. \{image\}\{question\}. The answer is: \{answer\}} as the instruction prompt.
We utilize public available datasets for supervised fine-tuning, including LLaVA-Mix-665K \cite{liu2023improved}, COCO Caption~\cite{chen2015cococaption}, VQAv2~\cite{goyal2017making}, ChartQA~\cite{masry2022chartqa}, DocVQA~\cite{clark2017docqa}, EST-VQA~\cite{wang2020estvqa}, InfoVQA~\cite{mathew2022infographicvqa}, STVQA~\cite{wang2020estvqa}, TextCaps~\cite{sidorov2020textcaps}, LLaVAR~\cite{zhang2023llavar}, OCR-VQA~\cite{mishra2019ocrvqa}, and DVQA~\cite{kafle2018dvqa}.
See \cref{tab:hp_vqa} for more training details.

\begin{table}[h]
    \centering
    \small
\resizebox{0.6\linewidth}{!}{
    \begin{tabular}{lc}
    \toprule
        Hyper-parameters & Value \\
    \midrule
        Input image resolution & $448\times 448$ \\

        \multirow{2}{*}{Learning rate} & 1e-6 (language model) \\
                                       & 1e-5 (others) \\
        Weight decay & 0.05 \\
        Warmup steps & 500 \\
        Learning rate schedule & cosine \\
        
        Training iterations & 10k \\
        
        Optimizer & AdamW \\
        Optimizer hyper-parameters & $\beta_1, \beta_2, \epsilon$ = 0.9, 0.999, 1e-8 \\
        
        Batch size & 256 \\
        
    \bottomrule
    \end{tabular}
}
    \vspace{0.3em}    
    \caption{Hyper-parameters for VQA and image captioning.}
    \vspace{-1em}
    \label{tab:hp_vqa}
\end{table}

\vspace{0.3em}
\noindent\textbf{Referring Expression Comprehension.}
Following previous works~\cite{chen2023shikra,bai2023qwenvl}, we train our model in the form of question answering, \ie, the prompt being \prompt{Question: Provide the bounding box coordinate of the region this sentence} 

\noindent{}\prompt{describes: \{object\}, Answer: (x1,y1)(x2,y2)}.
The generated bounding box is considered correct if its intersection over union (IoU) with the GT box is greater than 0.5.
Only public available datasets, including datasets from RefCOCO~\cite{kazemzadeh2014refcoco}, RefCOCO+~\cite{mao2016refcoco_plus_g}, and RefCOCOg~\cite{mao2016refcoco_plus_g} are utilized to train {\name}.
See \cref{tab:hp_reg} for fine-tuning hyper-parameters.

\begin{table}[t]
    \centering
    \small
\resizebox{0.6\linewidth}{!}{
    \begin{tabular}{lc}
    \toprule
        Hyper-parameters & Value \\
    \midrule
        Input image resolution & $224\times 224$ \\

        Learning rate & 2e-5 \\
        Weight decay & 0.05 \\
        Warmup steps & 500 \\
        Learning rate schedule & cosine \\
        
        Training iterations & 10k \\
        
        Optimizer & AdamW \\
        Optimizer hyper-parameters & $\beta_1, \beta_2, \epsilon$ = 0.9, 0.999, 1e-8 \\
        
        Augmentation & RandomHorizontalFlip \\
        Batch size & 256 \\
        
    \bottomrule
    \end{tabular}
}
    \vspace{0.3em}
    \caption{Hyper-parameters for referring expression comprehension.}
    \vspace{-2em}
    \label{tab:hp_reg}
\end{table}

\vspace{0.3em}
\noindent\textbf{Segmentation-to-Image Translation.}
~Following ControlNet~\cite{zhang2023adding}, we use BLIP~\cite{li2022blip} to generate text captions for each image in the ADE20K dataset. We train our model on the training set and evaluate on the validation set.
The input sequence is formulated as \prompt{\{segmentation image\}\{caption\}\{ground truth image\}} for each segmentation-image pair. 
See \cref{tab:hp_sft_img_gen} for more training details.

\begin{table}[t]
    \centering
    \small
\resizebox{0.6\linewidth}{!}{
    \begin{tabular}{lc}
    \toprule
        Hyper-parameters & ADE20K \\
    \midrule
        Input image resolution & $224\times 224$ \\
        Output image resolution & $512\times 512$ \\

        Learning rate & \makecell{1e-5 (image decoder) \\ 1e-4 (others)} \\
        Weight decay & 0.05 \\
        Warmup steps & 100 \\
        Learning rate schedule & cosine \\
        
        Training iterations & 4k \\
        
        Optimizer & AdamW \\
        Optimizer hyper-parameters & $\beta_1, \beta_2, \epsilon$ = 0.9, 0.98, 1e-5 \\
        
        Augmentation & RandomHorizontalFlip \\
        Batch size  & 512 \\
        
    \bottomrule
    \end{tabular}
}
    \vspace{0.3em}
    \caption{Hyper-parameters for Segmentation-to-Image Translation.}
    \vspace{-1em}
    \label{tab:hp_sft_img_gen}
\end{table}

\vspace{0.3em}
\noindent\textbf{Visual Storytelling.} Following previous works~\cite{koh2023gill,zheng2023minigpt5, pan2022ar_ldm}, \name{} is finetuned on the following three visual storytelling datasets respectively. The finetuning hyper-parameters are listed in \cref{tab:hp_sft_vist}.
\begin{itemize}
    \item \textbf{VIST}~\cite{huang2016vist} is a real-world vision-language dataset, containing 34k and 5k samples for training and evaluation. Each sample is a sequence consisting of 5 text captions and images. During training, we concatenate all the texts and images sequentially and the model is trained to predict all images. During inference, we test the model on generating the last image in the sequence, conditioned on all preceding images and texts following~\cite{koh2023gill,zheng2023minigpt5}. The evaluation metrics are FID~\cite{heusel2017gans} and the CLIP similarity~\cite{radford2021clip} between the generated images and the corresponding real images.
    
    \vspace{0.3em}
    \item \textbf{PororoSV}~\cite{li2019pororo} and \textbf{FlintstonesSV}~\cite{maharana2021integrating} are two cartoon storytelling datasets, containing 10191/2334/2208 and 20132/2071/2309 samples of the train, validation, and test set, respectively. Each sample is a sequence consisting of 5 text captions and frame images. During training, all the texts and images are concatenated sequentially and the model is trained to predict all images. During inference, the last 4 images are generated auto-regressively given the first image and all preceding captions as condition. FID~\cite{heusel2017gans} is used as the evaluation metric.
\end{itemize}

\begin{table}[t]
    \centering
    \small
\resizebox{0.7\linewidth}{!}{
    \begin{tabular}{lc}
    \toprule
        Hyper-parameters & VIST / PororoSV / FlintstonesSV \\
    \midrule
        Input image resolution & $224\times 224$ \\
        Output image resolution & $512\times 512$ \\

        Learning rate & \makecell{1e-4 (image decoder) \\ 1e-5 (others)} \\
        Weight decay & 0.05 \\
        Warmup steps & 200 \\
        Learning rate schedule & cosine \\
        
        Training iterations & 4k \\
        
        Optimizer & AdamW \\
        Optimizer hyper-parameters & $\beta_1, \beta_2, \epsilon$ = 0.9, 0.98, 1e-8 \\
        
        Augmentation & CenterCrop \\
        Batch size  & 128 \\
        
    \bottomrule
    \end{tabular}
}
    \vspace{0.3em}
    \caption{Hyper-parameters for visual storytelling.}
    \vspace{-2em}
    \label{tab:hp_sft_vist}
\end{table}

\vspace{-0.4em}
\subsection{Evaluation}

\begin{table*}[t]
\centering
\small
\resizebox{0.95\textwidth}{!}{
\begin{tabular}{@{}lllll@{}}
\toprule[0.95pt]
& Dataset & Task & Split & Metric \\
\cmidrule(l){2-5} 

\multirow{4}{*}{\rotatebox[origin=c]{90}{Caption.}} 

& COCO~\cite{chen2015cococaption}  & Scene description   & \texttt{test} &  CIDEr($\uparrow$)~\cite{CIDEr15} \\
& Flickr30k~\cite{plummer2015flickr30k}  & Scene description   & \texttt{test} &  CIDEr($\uparrow$)~\cite{CIDEr15} \\
& NoCaps~\cite{agrawal2019nocaps}  & Scene description   & \texttt{test} &  CIDEr($\uparrow$)~\cite{CIDEr15} \\
& Image2Paragraph~\cite{krause2017hierarchical}    & Scene description   & \texttt{test} &  CIDEr($\uparrow$)~\cite{CIDEr15} \\
\midrule[0.6pt]
\multirow{6}{*}{\rotatebox[origin=c]{90}{VQA.}} 
 & VQAv2~\cite{goyal2017making}        & Scene understanding QA & \texttt{test-dev}  &  VQA Acc($\uparrow$)~\cite{AntoVQA15}     \\
& OKVQA~\cite{marino2019ok}         & External knowledge QA &  \texttt{val}  &  VQA Acc($\uparrow$)~\cite{AntoVQA15}     \\
& GQA~\cite{hudson2019gqa}      & Scene understanding QA &  \texttt{test-dev}   &  VQA Acc($\uparrow$)~\cite{AntoVQA15}     \\
& VizWiz~\cite{gurari2018vizwiz}      & Scene understanding QA &  \texttt{test-dev}   &  VQA Acc($\uparrow$)~\cite{AntoVQA15}     \\
& TextVQA~\cite{singh2019towards}      & Text reading QA &  \texttt{val}   &  VQA Acc($\uparrow$)~\cite{AntoVQA15}     \\
& VisDial~\cite{das2017visual}      &  Image dialogue  &  \texttt{val}  & NDCG($\uparrow$)  \\

\midrule[0.6pt]
 \multirow{3}{*}{\rotatebox[origin=c]{90}{REC.}} 
& RefCOCO~\cite{kazemzadeh2014refcoco}      &  Referring experssion comprehension  &  -  & IoU Acc($\uparrow$)   \\
& RefCOCO+~\cite{mao2016refcoco_plus_g}      &  Referring experssion comprehension  &  -  & IoU Acc($\uparrow$)   \\
& RefCOCOg~\cite{mao2016refcoco_plus_g}      &  Referring experssion comprehension  &  -  & IoU Acc($\uparrow$)   \\
 
\midrule[0.6pt]
\multirow{6}{*}{\rotatebox[origin=c]{90}{Generation.}} 
 & MS-COCO~\cite{lin2014microsoft} & Text-to-image generation  &  \texttt{val-30K}   & FID($\downarrow$)~\cite{heusel2017gans}\\
 & LN-COCO~\cite{pont2020connecting} & Text-to-image generation  & \texttt{val} & FID($\downarrow$)~\cite{heusel2017gans}\\
& ADE20k~\cite{zhou2017ade20k} & Segmentation-to-image generation  & \texttt{val} & mIoU($\uparrow$)\\
& VIST~\cite{huang2016vist} & Interleaved-context image generation  & \texttt{val} & CLIP-Sim($\uparrow$)~\cite{radford2021clip}, FID($\downarrow$)~\cite{heusel2017gans}\\
& PororoSV~\cite{li2019pororo} & Interleaved-context multi-image generation  & \texttt{test} & FID($\downarrow$)~\cite{heusel2017gans}\\
& FlintstonesSV~\cite{gupta2018flintstones} & Interleaved-context multi-image generation  & \texttt{test} & FID($\downarrow$)~\cite{heusel2017gans}\\

\bottomrule[0.95pt]

\end{tabular}
}
\vspace{0.3em}
\caption{\textbf{Summary of evaluation benchmarks,} including image caption, visual question answering, referring experssion comprehension, and image generation.}
\vspace{-2em}
\label{tab:supp_eval}

\end{table*}

\noindent\textbf{Benchmarks}. Evaluating \name{} comprehensively requires various benchmarks and datasets, such as image caption, visual question answering, text-to-image generation and so on. All these evaluation tasks and metrics are listed in \cref{tab:supp_eval}.

\vspace{0.3em}
\noindent\textbf{Image Generation}. For all image generation tasks, the scale of classifier-free guidance~\cite{ho2022classifier} and the total inference step is set as 3.5 and 250 by default.

\vspace{0.3em}
\noindent\textbf{Text Generation}. The prompt templates for each text generation tasks are listed in \cref{table:supp_prompt}. The `Image Caption (short)' task includes COCO Caption~\cite{chen2015cococaption}, Flicker30k~\cite{plummer2015flickr30k} and  NoCaps~\cite{agrawal2019nocaps}, while the `Image Caption (long)' task includes Image2Paragraph~\cite{krause2017hierarchical}.

\begin{table*}[t]
\small
\centering
\renewcommand{\arraystretch}{2.0}
\resizebox{0.9\linewidth}{!}{
\begin{tabular}{@{}cm{3.5cm}|m{10cm}@{}}
\toprule

  & Task  & Prompt Template \\ \midrule

\multirow{7}{*}{Zero-shot}  &   \multirow{1}{*}{Image Caption (short)} & \prompt{\{image\} a photo of}           \\
                            &   \multirow{1}{*}{Image Caption (long)}  & \prompt{\{image\} Please describe the image in detail. The image depicts}           \\
                             &   \multirow{1}{*}{VQA (except VizWiz)}  & \prompt{Based on the image, please answer the question.  \{image\}\{question\}  Please provide an accurate answer within one word. The answer is:}           \\
                             &   \multirow{1}{*}{VizWiz QA}  & \prompt{Based on the image, please answer the question. \{image\}\{question\}  When the provided information is insufficient, respond with 'Unanswerable'. Please provide an accurate answer within one word. The answer is:}          \\
                             &   \multirow{1}{*}{Visual Dialog}  & \prompt{\{image\} caption: \{caption\} question: \{history question\}? answer: \{history answer\}.  $\cdots$ question: \{question\}? answer:}           \\
 \midrule

 \multirowcell{7}{Supervised \\ Fine-tuning}  &   Image Caption (short) & \prompt{\{image\} Provide a one-sentence caption for the provided image.}              \\

                             &   Image Caption (long)  & \prompt{\{image\} Please describe the image in detail.}          \\
                             &   VQA (except VizWiz)  & \prompt{Based on the image, please answer the question.  \{image\}\{question\}  Please provide an accurate answer within one word. The answer is:}          \\
                             &   VizWiz VQA  & \prompt{Based on the image, please answer the question. \{image\}\{question\}  When the provided information is insufficient, respond with 'Unanswerable'. Please provide an accurate answer within one word. The answer is:}           \\
                             &   Visual Dialog  & \prompt{\{image\} caption: \{caption\} question: \{history question\}? answer: \{history answer\}.  $\cdots$ question: \{question\}? answer:}           \\

\bottomrule
\end{tabular}
}
\vspace{0.3em}
\caption{Prompt templates for text generation.}
\vspace{-2em}
\label{table:supp_prompt}
\end{table*}

\section{Additional Ablation Studies}

This section provides more ablation studies for \name{}, all of which share the same settings as those in the main text by default. Note that these models are not fine-tuned on downstream tasks.

\vspace{0.5em}\noindent{\textbf{Pre-training with Different Loss Terms}}. As shown in \cref{table:supp_loss}, \name{} achieves better performance when jointly trained with both $\mathcal{L}_{NTP}$ and $\mathcal{L}_{NIP}$  on comprehension and generation tasks, demonstrating the mutual benefits between the two loss terms. Moreover, setting $\lambda=10$ achieves a better balance between $\mathcal{L}_{NTP}$ and $\mathcal{L}_{NIP}$ empirically.

\begin{table}[h]
\footnotesize
\centering
\setlength\tabcolsep{2pt}
\resizebox{0.7\linewidth}{!}{
\begin{tabular}{@{}l|cccc@{}}
\toprule

 Loss Term  & Caption$\uparrow$ & Generation$\downarrow$ & OK-VQA$\uparrow$ & TextVQA$\uparrow$ \\ \midrule

$\mathcal{L}_{NTP} + 100~\mathcal{L}_{NIP}$    &   106.2 & 31.1 & \textbf{29.8} & 24.5 \\
\rowcolor{mygray} 
$\mathcal{L}_{NTP} + 10~\mathcal{L}_{NIP}$    &   \textbf{110.6}       & \textbf{30.0}      &  \textbf{29.8}  &   \textbf{27.7} \\

$\mathcal{L}_{NTP} + \mathcal{L}_{NIP}$    &   \textbf{110.0} & 31.4 & 29.3 & 26.0 \\

$\mathcal{L}_{NTP} $  only     &    105.7      &  --     & \textbf{29.9}  &  \textbf{27.6}  \\

$\mathcal{L}_{NIP}$  only  &    --      &  34.2     & --  &  --  \\

\bottomrule
\end{tabular}
}
\vspace{0.5em}
\caption{\textbf{Pre-training with different loss terms.} }
\vspace{-1.5em}
\label{table:supp_loss}
\end{table}

\vspace{0.3em}
\noindent{\textbf{The Relationship between MMFS and Resampler}}. The results in \cref{table:supp_resampler} validate the mutual benefits between our proposed MMFS module and the Resampler used in image tokenizer, as removing either of them leads to an overall performance degradation. 

\begin{table}[h]
\footnotesize
\centering
\setlength\tabcolsep{2pt}
\resizebox{0.75\linewidth}{!}{
\begin{tabular}{@{}cc|cccc@{}}
\toprule

 w/ Resampler & w/ MMFS  & Caption$\uparrow$ & Generation$\downarrow$ & OK-VQA$\uparrow$ & TextVQA$\uparrow$ \\ \midrule
\rowcolor{mygray} 
     \ding{51}   &     \ding{51}   &   \textbf{110.6}       & \textbf{30.0}      &  \textbf{29.8}  &   \textbf{27.7} \\

     \ding{51}        &               &   107.0       &  32.2     & 28.7   & 22.5   \\

             &      \ding{51}         &      102.7    &   32.0    & 27.3  &  22.0  \\

\bottomrule
\end{tabular}
}
\vspace{0.5em}
\caption{\textbf{The complementary relationship between MMFS and Resampler.} When not using Resampler, we directly feed 32 randomly-initialized learnable embeddings as input visual tokens into the LLM.}
\vspace{-2em}
\label{table:supp_resampler}
\end{table}

\section{Qualitative Results}

\subsection{Zero-shot Results}
\begin{figure*}[ht]
    \centering
    \includegraphics[width=0.95\linewidth]{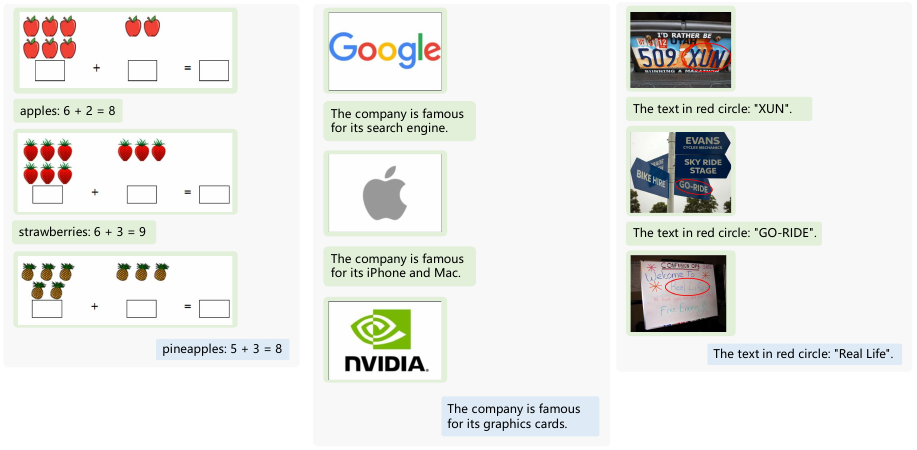}
    \caption{
        \textbf{Zero-shot text generation with interleaved images and texts} serving as multi-modal few-shot contexts. 
    }
    \label{fig:vis_zero_shot_3}
\end{figure*}

\begin{figure*}[ht]
    \centering
    \includegraphics[width=0.75\linewidth]{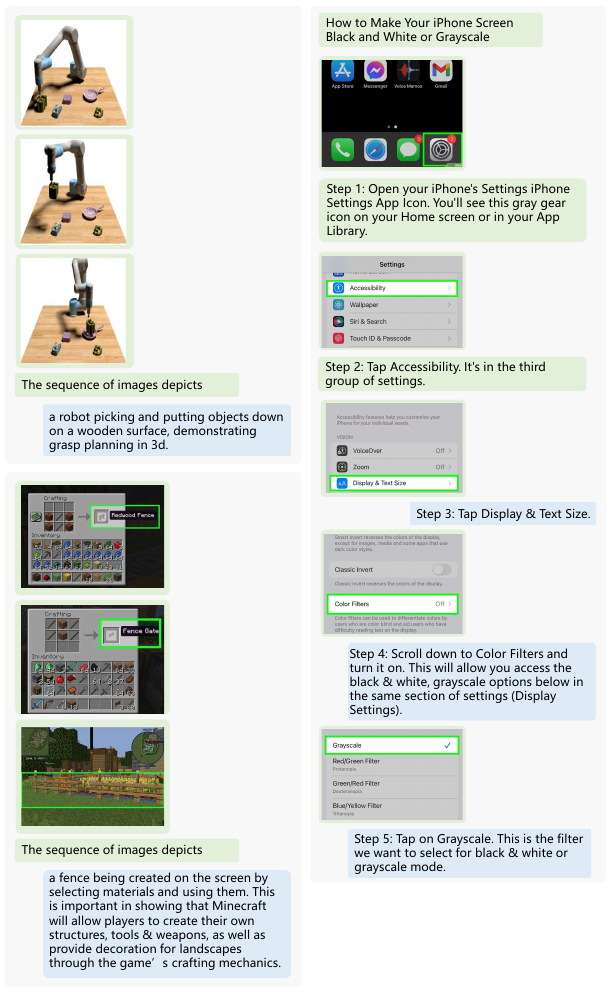}
    \caption{
        \textbf{Zero-shot text generation with interleaved images and texts} on complex scenarios such as robotics, gaming and GUI interface. 
    }
    \label{fig:vis_zero_shot_2}
\end{figure*}

\begin{figure*}[ht]
    \centering
    \includegraphics[width=0.75\linewidth]{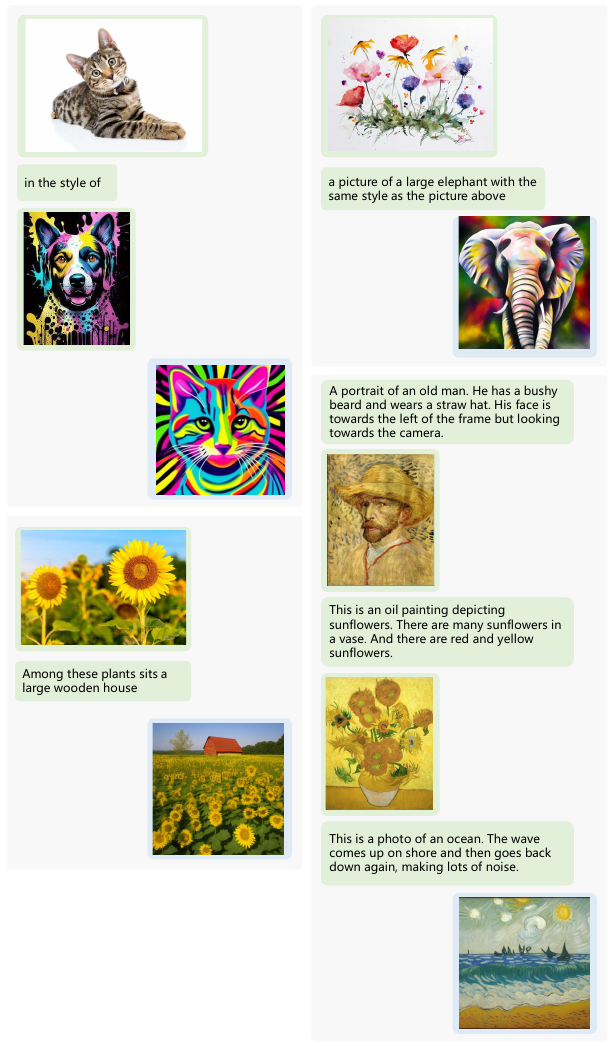}
    \caption{
        \textbf{Zero-shot image generation with interleaved images and texts}. Our method can generate appropriate images based on the provided contexts of desired styles or concepts.
    }
    \label{fig:vis_zero_shot_1}
\end{figure*}

\begin{figure*}[ht]
    \centering
    \includegraphics[width=0.98\linewidth]{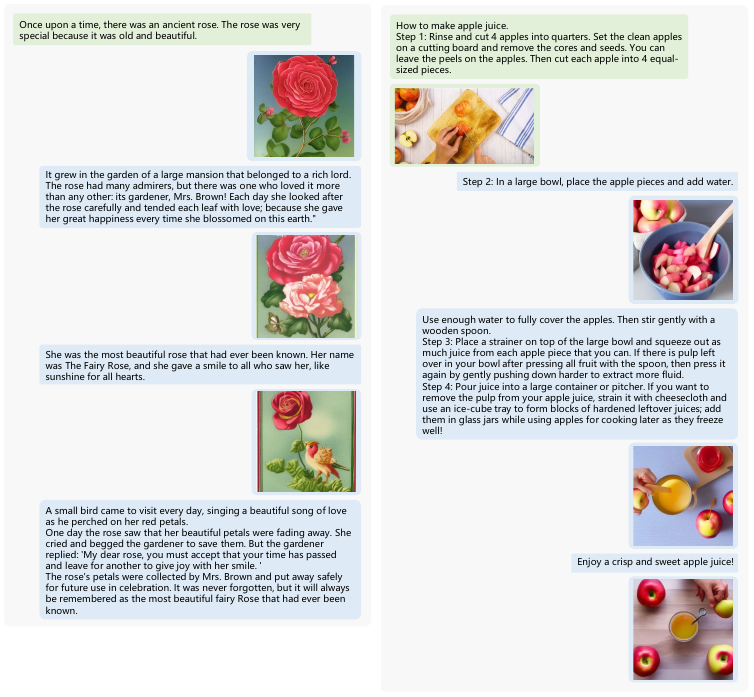}
    \caption{
        \textbf{Zero-shot interleaved image and text generation} for visual storytelling and multi-modal instructions. 
    }
    \label{fig:vis_zero_shot_4}
\end{figure*}

The zero-shot comprehension and generation visualizations of our method are shown in \cref{fig:vis_zero_shot_3,fig:vis_zero_shot_2,fig:vis_zero_shot_1,fig:vis_zero_shot_4}. Specifically, \cref{fig:vis_zero_shot_3} demonstrates the emergent multi-modal in-context learning capability of our method, while \cref{fig:vis_zero_shot_2} further illustrates the effectiveness of our method in understanding complex scenarios such as robotics, gaming and GUI interface. \cref{fig:vis_zero_shot_1} exhibits that our method can also generate appropriate images based on the provided contexts of desired styles or concepts. \cref{fig:vis_zero_shot_4} displays our method's capability of generating images and texts interleavedly.

\subsection{Supervised Finetuning Results}

\begin{figure*}[ht]
    \centering
    \includegraphics[width=\linewidth]{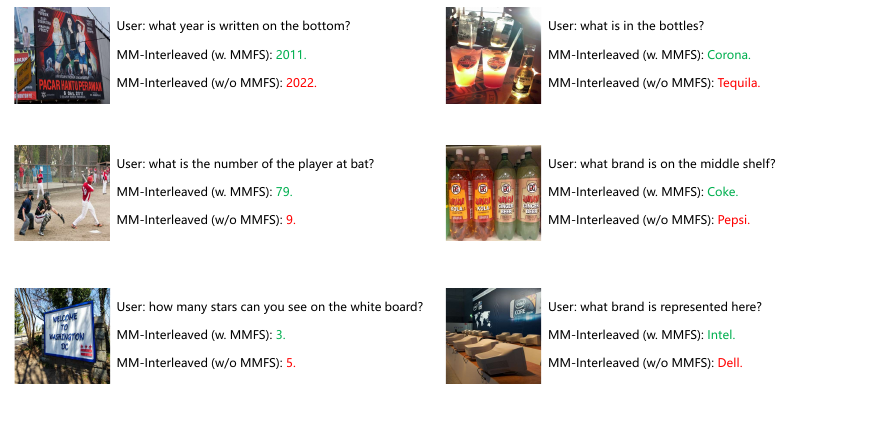}
    \caption{
        \textbf{Qualitative results on TextVQA~\cite{singh2019towards}}. 
        Each example consists of the user query, the answer given by \name{} with \textit{MMFS}, and the answer given by \name{} without \textit{MMFS}. The image shapes are normalized for visualization.
    }
    \label{fig:textvqa}
\end{figure*}

\noindent\textbf{Text Reading QA}. As is shown in \cref{fig:textvqa}, \name{} with MMFS provides more accurate answers when requiring fine-grained details for generating text outputs given VL inputs.

\begin{figure*}[ht]
    \centering
    \includegraphics[width=\linewidth]{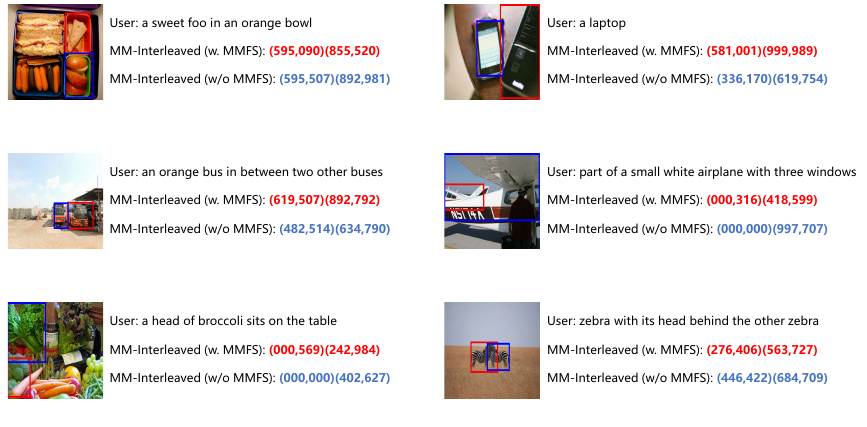}
    \caption{
        \textbf{Referring Expression Comprehension on RefCOCOg~\cite{mao2016refcoco_plus_g}}. 
        Each example consists of the user query, the box predicted with \textit{MMFS}, and the box predicted without \textit{MMFS}. The image shapes are normalized for visualization.
    }
    \vspace{-1.5em}
    \label{fig:grounding}
\end{figure*}

\vspace{0.3em}
\noindent\textbf{Referring Expression Comprehension}. The visualization of \name{} on REC tasks is shown in \cref{fig:grounding}. Our model with MMFS is also capable of generating more accurate coordinates given the referring expression and the query image. 

\begin{figure*}[ht]
    \centering
    \includegraphics[width=0.55\linewidth]{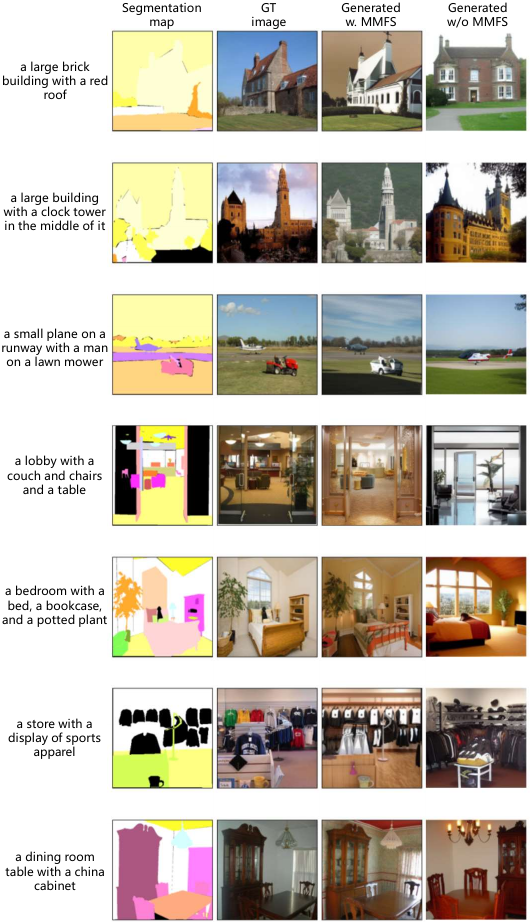}
    \caption{
        \textbf{Segmentation-to-Image Generation on ADE20k~\cite{zhou2017ade20k}}. 
        Each row is an example consisting of four images, which are the input segmentation map, the ground truth image, the generated image \textit{with MMFS}, and the generated image  \textit{without MMFS} respectively. The shape of ground-truth images and segmentation maps is normalized for visualization. When \textit{without MMFS}, the generated results lack spatial alignment with the input segmentation maps. 
    }
    \vspace{-1.5em}
    \label{fig:segm}
\end{figure*}

\vspace{0.3em}
\noindent\textbf{Segmentation-to-image Translation}. \cref{fig:segm} shows the visualization results of \name{} for segmentation-to-image translation. Given the text prompt and segmentation map, the spatial layout of images generated with MMFS is significantly closer to the original ground-truth image, compared to the baseline results without MMFS.  

\begin{figure*}[ht]
    \centering
    \includegraphics[width=0.95\linewidth]{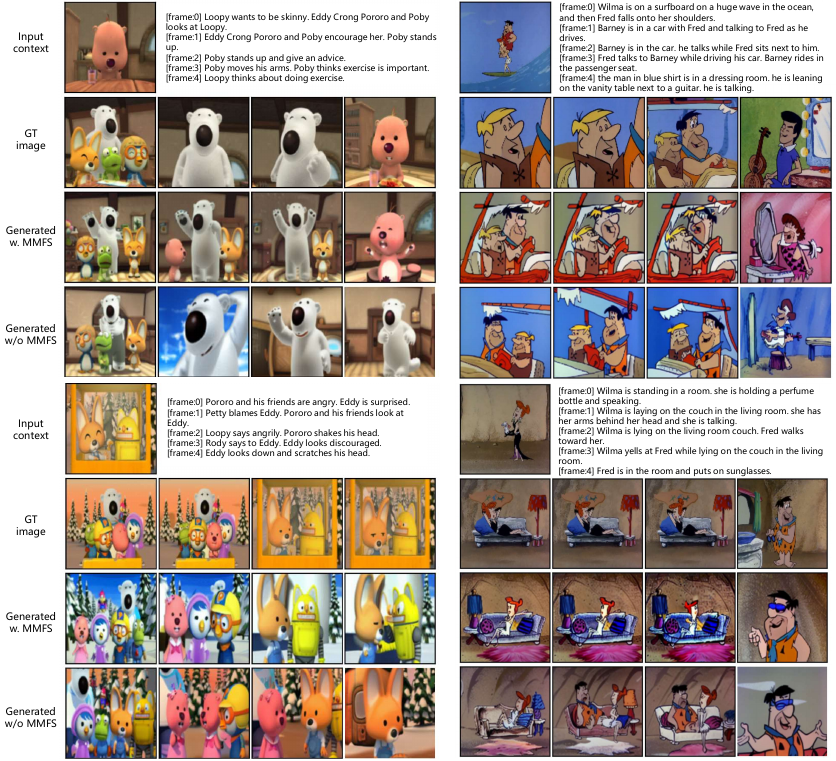}
    \caption{
        \textbf{Multi-image Generation on PororoSV~\cite{li2019pororo} and FlintstonesSV~\cite{gupta2018flintstones}}. 
        Each example consists of four rows. The first row is the first frame image and all corresponding captions. The second row contains the ground truth images of following frames; The third row are the generated results \textit{with MMFS}; And the last row are the results generated \textit{without MMFS}. When \textit{without MMFS}, the generated multiple images lack content consistency in terms of characters, backgrounds, objects, etc.
    }
    \vspace{-1.5em}
    \label{fig:vist}
\end{figure*}

\vspace{0.3em}
\noindent\textbf{Multi-image Generation}. In \cref{fig:vist}, we compare the multiple images sequentially generated by \name{} with and without MMFS. The images generated with MMFS achieves better spatial consistency (\eg the background environment, change of viewpoint, character position relationship \textit{etc.}) and closer semantic alignment with the interleaved image-text context. 

\begin{figure*}[ht]
    \centering
    \includegraphics[width=0.88\linewidth]{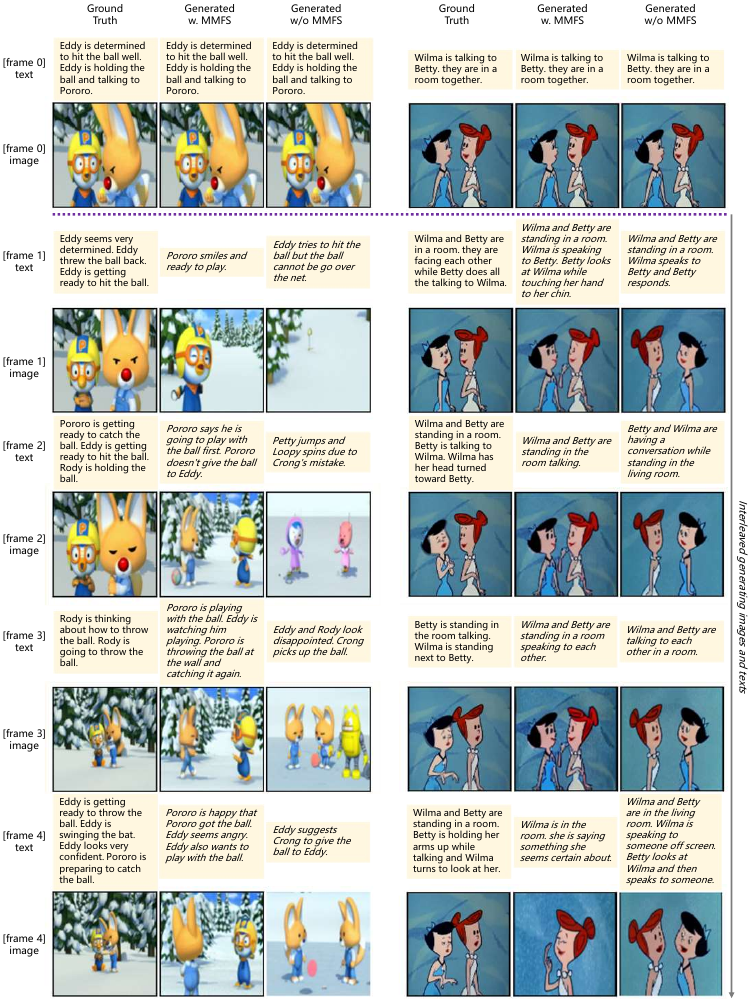}
    \caption{
        \textbf{Interleaved Image-Text Generation on PororoSV~\cite{li2019pororo} and FlintstonesSV~\cite{gupta2018flintstones}}. 
        Each example consists of three columns. The first column is the ground-truth images and captions of all frames. The second column are the generated results \textit{with MMFS}; And the last column are the results generated \textit{without MMFS}. Only the caption and image of the first frame is given as condition during generation. 
    }
    \vspace{-1.5em}
    \label{fig:vist_interleaved}
\end{figure*}

\vspace{0.3em}
\noindent\textbf{Generating Interleaved Image and Texts.} Moreover, the model is further finetuned to generate both images and texts simultaneously for visual storytelling tasks. As is shown in \cref{fig:vist_interleaved}, given the first frame image and caption as context, \name{} with MMFS generates the following interleaved images and texts coherently, achieving balance between the generation diversity and spatial semantic consistency.

\end{document}